\documentclass[letterpaper, 10 pt, journal, twoside]{IEEEtran}

\usepackage{amsmath} % assumes amsmath package installed
\usepackage{amssymb}  % assumes amsmath package installed
\usepackage[protrusion=false]{microtype}
\usepackage{tensor}
\usepackage{mathtools}
\usepackage{makecell}
\usepackage[colorinlistoftodos,prependcaption]{todonotes}
\usepackage{siunitx}
\usepackage{comment}
\usepackage{xcolor} % for textcolor
\usepackage[normalem]{ulem}
\usepackage{amsmath}
\usepackage{amsfonts}
\usepackage{amssymb}
\usepackage{float}
\usepackage{lipsum}
\usepackage{multicol}
\usepackage{multirow}
\usepackage{caption} %For sub-figures
\usepackage{subcaption} %For sub-figures

\usepackage[linesnumbered,ruled,vlined]{algorithm2e}

\usepackage{booktabs}

\title{\LARGE \bf
	IMU-based Online Multi-lidar Calibration}

\author{Sandipan Das$^{1,2}$, Bengt Boberg$^2$, Maurice Fallon$^3$, Saikat Chatterjee$^{1}$
	\thanks{$^1$ KTH EECS, Sweden. \texttt{\{sandipan,sach\}@kth.se}\newline%
			$^2$ Scania, Sweden. \texttt{\{sandipan.das, bengt.boberg\}@scania.com}\newline%
			$^3$ ORI, University of Oxford, UK. \texttt{mfallon@robots.ox.ac.uk}\newline%
		 	$^4$ \url{https://github.com/mrsandipandas/imu-calibration}}}

\usepackage[linesnumbered,ruled,vlined]{algorithm2e}

\usepackage{multirow}
\usepackage[normalem]{ulem}
\usepackage{xargs} 

\newcommand{\hide}[1]{}
\newcommand{\Figure}{Fig.~}

\newcommand{\bdmath}{\begin{dmath}}
\newcommand{\edmath}{\end{dmath}}
\newcommand{\beq}{\begin{equation}}
\newcommand{\eeq}{\end{equation}}
\newcommand{\bdm}{\begin{displaymath}}
\newcommand{\edm}{\end{displaymath}}
\newcommand{\bea}{\begin{eqnarray}}
\newcommand{\eea}{\end{eqnarray}}
\newcommand{\beal}{\beq \begin{array}{ll}}
\newcommand{\eeal}{\end{array} \eeq}
\newcommand{\beas}{\begin{eqnarray*}}
\newcommand{\eeas}{\end{eqnarray*}}
\newcommand{\ba}{\begin{array}}
\newcommand{\ea}{\end{array}}
\newcommand{\bit}{\begin{itemize}}
\newcommand{\eit}{\end{itemize}}
\newcommand{\ben}{\begin{enumerate}}
\newcommand{\een}{\end{enumerate}}

\newcommand{\SO}{\mathrm{SO}}

\newcommand{\Real}{\mathbb{R}}

\newcommand{\SEthree}{\ensuremath{\mathrm{SE}(3)}\xspace}

\newcommand{\SOthree}{\ensuremath{\SO(3)}\xspace}

\newcommand{\calF}{{\cal F}}

\newcommand{\T}{\mathbf{T}}
\newcommand{\R}{\mathbf{R}}

\newcommand{\Identity}{\mathbf{I}}

\newcommand{\eye}{{\mathbf I}}

\newcommand{\tran}{\mathbf{t}}

\newcommand{\eq}{Eq.}

\newcommand{\World}{\mathtt{W}}
\newcommand{\Imu}{\mathtt{I}}
\newcommand{\Camera}{\mathtt{C}}
\newcommand{\Lidar}{\mathtt{L}}

\newcommand{\world}{\mathtt{{W}}}

\newcommand{\imu}{\mathtt{{I}}}

\newcommand{\Base}{\mathtt{{B}}}

\SetCommentSty{algo}
\SetKwInput{KwInput}{Input} % Set the Input
\SetKwInput{KwOutput}{Output} % set the Output

\DeclareMathOperator*{\argmax}{arg\,max}
\DeclareMathOperator*{\argmin}{arg\,min}

\newcommand\cancel{\bgroup\markoverwith{\textcolor{red}{\rule[0.5ex]{2pt}{0.4pt}}}\ULon}

\makeatletter
\let\NAT@parse\undefined
\makeatother
\usepackage{hyperref}
\begin{document}
	
	\maketitle
	\thispagestyle{empty}
	\pagestyle{empty}
	
	Modern autonomous systems typically use several sensors for perception. For best performance, accurate and reliable extrinsic calibration is necessary. In this research, we propose a reliable technique for the extrinsic calibration of several lidars on a vehicle without the need for odometry estimation or fiducial markers. First, our method generates an initial guess of the extrinsics by matching the raw signals of IMUs co-located with each lidar. This initial guess is then used in ICP and point cloud feature matching which refines and verifies this estimate. Furthermore, we can use observability criteria to choose a subset of the IMU measurements that have the highest mutual information --- rather than comparing all the readings. We have successfully validated our methodology using data gathered from Scania test vehicles.

	%\mfallon{you cannot put a Github and the link in the abstract. The abstract is used by IEEE and the abstract looks silly on its own with this footnote number. You can put it into the end of the Contributions instead.}

	% Keep Consistent Tenses:
	%in background/lit review: past
	%for experiments carried out: past
	%for results being evaluated currenty: present
	
    %\mfallon{you need to include a quantitative claim here - `improved by X percent' or `estimated the calibration to within Xmm as verified from CAD'. AFAI can tell you dont have a baseline algorithm to compare to. Can you add one?}
	
	\section{Introduction}
	\label{sec:introduction}
	
	For safe navigation and sensor redundancy, autonomous vehicles need numerous sensors to ensure $360^{\circ}$ sensing coverage. For precise fusion, it is crucial to establish the accurate mounting location for each of the vehicle's sensors --- a process known as extrinsic calibration. Without the proper extrinsic calibration parameters, it is impossible to fuse the sensing data into a single reference frame. In this work, we concentrated on multi-lidar system extrinsic calibration, though the same principle might also be applied to camera or radar extrinsic calibration.
	
	There are two main types of multi-lidar extrinsic calibration algorithms: target-based and target-less. Target-based calibration refers to putting fiducial markers in the shared field-of-view (FoV) of the lidars and mutually matching the markers. Target-less calibration is carried out by comparing the estimated states of the sensors for initialization and refined using feature-matching. In another set of target-less methods, both state and the calibration parameters are estimated by solving a joint optimization step where the reprojection error between the tracked features is minimized. Since the sensors may not have an overlapping FoV, target-less calibration is preferable as it is more general than target-based methods. It is widely recognized that not all motion segments produce enough observable information for matching states used for the initialization of extrinsic calibration. Because of this identifying any degenerate motion segments is also important to discard information that is not helpful for extrinsic calibration initialization.

	%\mfallon{I'm not keen on citing blocks of papers like this. It doesnt help - as there is no narrative. I'd recommend removing some of these especially if they appear in the Related Work}

    %\mfallon{`alternative' has to be related to another method. So I suggest just saying `novel'}	
	%In this paper, we present a novel vehicle-run, observability-aware, target-less calibration method for simultaneously calibrating multiple lidars using co-located IMU (Inertial measurement unit) signals without the need for state estimation. % Additionally, we compensate for the biases of the IMU measurements online to improve the calibration accuracy.
	%\mfallon{this previous paragraph is not necessary - you have the contribution subsection to come yet! Repeation}
	
	\begin{figure}
		\centering
		\includegraphics[width=1\columnwidth]{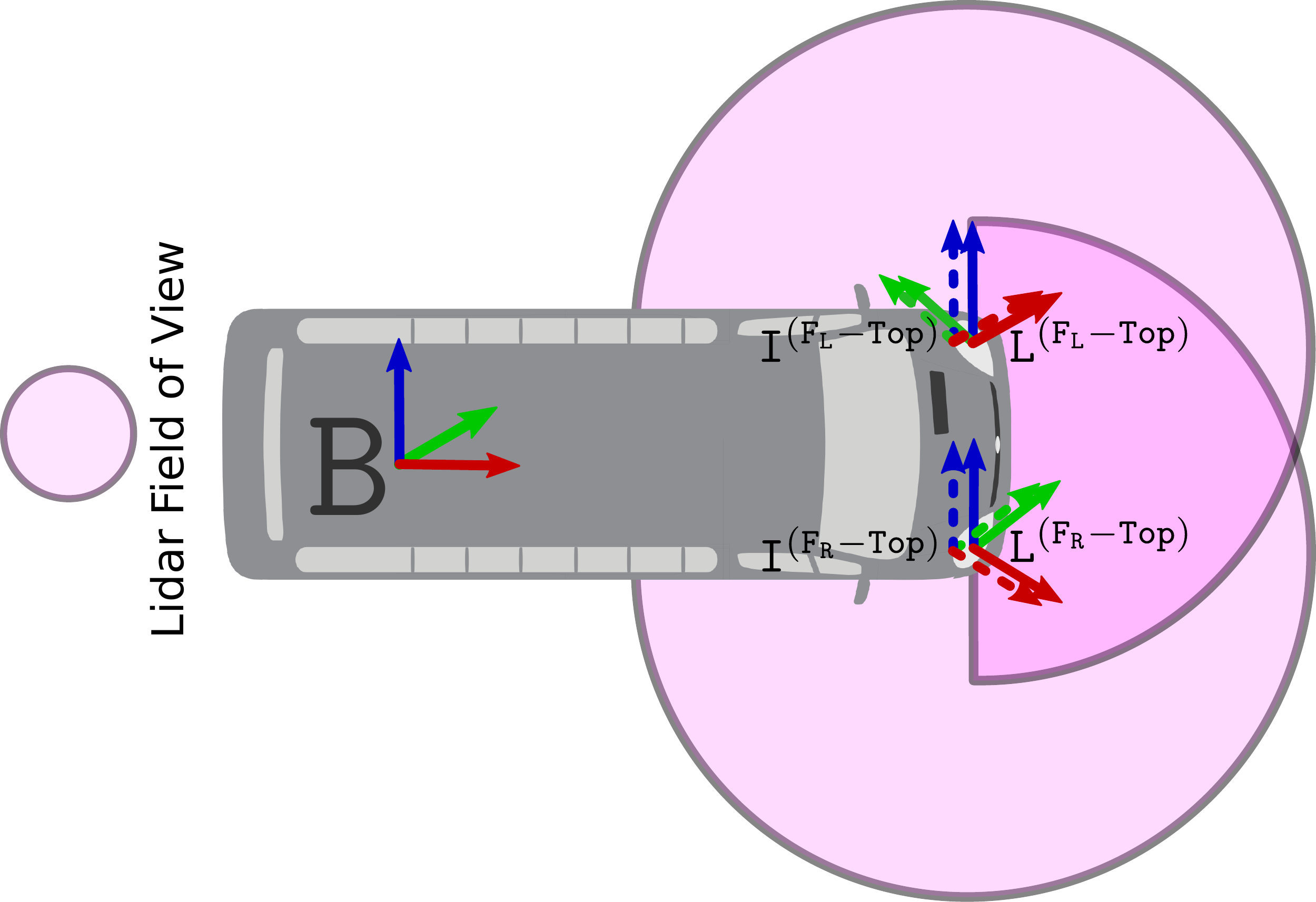}
		%\vspace{-5mm}
		\caption{Illustration of 2 lidars with co-located IMUs positioned
			at the corners of our data collection vehicle. The vehicle base frame $\Base$, is located
			at the center of the rear axle. The sensor frames of the lidars (solid lines) are:
			$\Lidar^{(\mathtt{F}_\mathtt{L}-\mathtt{Top})}$ and $\Lidar^{(\mathtt{F}_\mathtt{R}-\mathtt{Top})}$ whereas, the sensor frames of the IMUs (dotted lines) are:
			$\Imu^{(\mathtt{F}_\mathtt{L}-\mathtt{Top})}$ and $\Imu^{(\mathtt{F}_\mathtt{R}-\mathtt{Top})}$.
			${(\mathtt{F}_\mathtt{L})}$: front-left and ${(\mathtt{F}_\mathtt{R})}$: front-right.} 
		\label{fig:scania-bevda}
		\vspace{-5mm}
	\end{figure}
	
	\subsection{Motivation}
	As seen in \Figure\ref{fig:scania-bevda} and \Figure\ref{fig:coordinate-frames}, the mobile platforms we employed for our studies have multiple lidars. Therefore, it is essential to confirm that the sensors are correctly calibrated prior to any autonomous run. To properly excite the different degrees of freedom, motion-based calibration requires aggressive driving maneuvers which can cause state estimation drift.
	%The state estimates for motion-based calibration have intrinsic drifts under aggressive maneuvers required for target-less calibration to excite the different degrees of freedom. \mfallon{previous sentence is not a logical argument. what about: `'}
	Moreover, state estimation itself adds additional complexity to the whole tool chain and it does not provide independent verification for the estimated extrinsics. Finally, we also need a process to verify the quality of the estimated extrinsics.
	
	%\mfallon{I'm not certain but I think it should be co-located not co-located. I think the latter means something else.}
	%This is why we propose to initialize the lidar extrinsics by matching the raw IMU signals (with bias compensation) between the co-located IMU(s) present with the lidar(s) without the need for any rotation initialization. Afterward, for extrinsics refinement we match line features between the lidars using the obtained initialization. The approach can also provide online verification of the estimate by matching plane features. Furthermore, we solely compare the raw angular velocity data for observability analysis.

	%\mfallon{nothing in the previous paragraph is motivation - its contribution ... and I can see its repeated in the contributions section.}
	
	\subsection{Contribution}
	\label{sec:contribution}
	%\mfallon{the next sentence is silly - remove}
	We present the following contributions:
	\begin{itemize}
		\item A novel observability-aware, target-less extrinsic calibration algorithm to calibrate multiple lidars using co-located IMUs without the need for any lidar odometry estimation and extrinsics rotation initialization.
		\item Online extrinsics refinement based on line feature matching and online bias compensation of the IMU signals to improve the quality of the extrinsic calibration algorithm. %\sandipan{with specific data collection pattern?}
		\item Verification of our method in real-time based on plane feature association with datasets collected from Scania test vehicles in urban driving scenarios.
		\item Our implementation for the IMU-IMU calibration with observability analysis on a custom-made test board and sample datasets is open-sourced on Github$^4$.
	\end{itemize}
	
	\section{Related Work}
	\label{sec:related-work}	
	Extrinsic sensor calibration is a well studied area. In our discussion, we briefly review the relevant literature and motivate our choice of methods.
	
	\subsection{Target-based calibration}
	Target-based extrinsic calibration is performed by matching known markers or fiducials placed in the common FoV of multiple sensors. Researchers have explored target-based between camera and lidar using known geometric shapes like checkerboards, triangles, and diamonds \cite{huang2020lidarcam} or geometric feature correspondences \cite{zhou2018auto}. To calibrate a multi-lidar system, geometric elements including points, lines, and planes were retrieved and matched in \cite{He2013, Choi2016}. The most important aspect of a target-based system is there being overlapping FoV.
	
	\subsection{Target-less calibration}
	Target-less extrinsic calibration based on the Hand-Eye method \cite{brookshire2013extrinsic,hand-eye_tsai_1989,horaud1995hand} is an extensively studied topic. The term ``Hand-Eye" refers to early motion-based calibration research that estimated the motion of the gripper (hand) and the camera (eye) while constricting their poses with a fixed rigid body transformation. Multi-camera extrinsic calibration with Hand-Eye method has been explored in works like Camodocal \cite{heng2013camodocal} and recently extended to multi-lidar extrinsic calibration \cite{jiao2021robust}.

	Another technique for target-less calibration is pose alignment with Kabsch method \cite{Kabsch1978}, where the poses are obtained from lidar odometry \cite{shan-legoloam, xu2022fast} in the corresponding sensor frame. Recent works like CROON \cite{wei2022croon} and adaptive voxelization \cite{liu2022adaptive} improve the pose alignment with additional feature matching techniques. In Kalibr \cite{kalibr_meye_2013} the authors automatically detected sets of measurements from which they could identify an observable parameter space and then performed a maximum likelihood estimate (MLE) by minimizing the errors
	between landmark observations and their known correspondences. They also discarded parameter updates for numerically unobserved directions and degenerate scenarios. In \cite{tu2022lidarsfmcalib}, a global bundle
	adjustment was performed to minimize the reprojection error between the tracked features to estimate the camera-lidar extrinsics. A similar principle can be applied for multiple lidar extrinsic calibration if the sensors have an overlapping FoV. However, this method would still require lidar odometry estimates, which is a computationally expensive process.

	%\mfallon{Are you sure about the above? We use Kalibr for checkerboard based calibration}
	%\sandipan{Yes, they do this in Kalibr. The location of markers in chekerboard is the known correspondance, which they use!}
	
	Because we have integrated IMUs within our lidars (with factory calibrations), we aim to use them to create a multi-IMU (MIMU) configuration to estimate initial extrinsics estimates without the need for any pose alignment. Kalibr was extended towards MIMU calibration \cite{rehder2016extending} by matching poses fitted to a B-spline and further constraining motion with known image landmarks. In MIMC-VINS \cite{mimc_vins}, an efficient multi-state constraint Kalman filter was used to jointly propagate all IMU states while enforcing rigid body constraints between the IMUs during the filter update stage, which produced the MIMU calibration.
	Unlike prior works, we used the fundamental property that a rigid body's angular velocity is constant around every point of the body to formulate our MIMU calibration. We used MLE to match the raw angular velocity between the IMU(s) embedded within the lidar(s) in accordance with the signal-to-signal match principle \cite{pittelkau2005calibration}.

	We then selected measurements that have contain sufficient signal excitation which is known as observability-aware criteria \cite{hausman2017observability, schneider2019observability, lv2022observability, das2023observability}. We then maximized mutual information between the angular velocity signals to identify relevant motion segments. %We also estimated the IMU biases online by capturing data in a special sequence, removing the need for any state estimation for extrinsics initialization. Finally, we performed point cloud feature extraction and matching for extrinsics refinement and verification in real time. 
	This method can be extended to other modalities and does not require overlapping FoV between the sensors.

	%\mfallon{the previous paragraph is not related work - its a discussion of your method and really part of the problem statement}
	
	%\sandipan{CROON is single frame. Adaptive voxel suitable for static scenes. With Kabsch alignment we may have bad state estimates. Both CROON and adaptive voxels are computationally heavy. IMO adaptive voxel is meant for indoor feature rich environment. For refinement they insist the platform must not move. Other methods, translation and rotational initialization needed.}
	
	\section{Problem Statement}
	\label{sec:problem-statement}
	
	\subsection{Sensor platform and reference frames}
	\label{sec:sensor-platform}
	\begin{figure}
		\centering
		\vspace{2mm}
		\includegraphics[width=1\columnwidth]{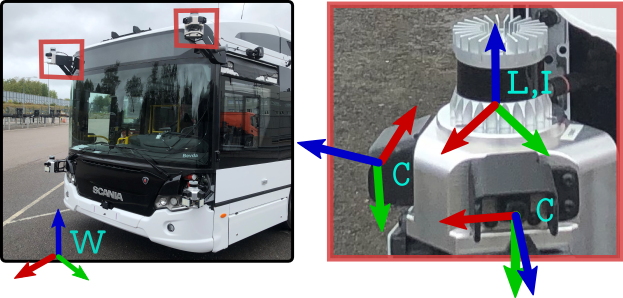} 
		\vspace{-8mm}
		\caption{Reference frames conventions for our vehicle platform. The world frame
			$\World$ is a fixed frame, while the base frame $\Base$, as shown in \Figure
			\ref{fig:scania-bevda}, is located at the rear axle center of the vehicle.
			Each
			sensor unit contains the two optical frames $\Camera$, an IMU frame, $\Imu$,
			and
			a lidar frame $\Lidar$.} %\sandipan{I think we can improve this figure by
		%adding
		%a schematic of frames, in ros we can model the robot by urdf model}
		\label{fig:coordinate-frames}
		\vspace{-5mm}
	\end{figure}
	We evaluate our proposed method using data collected from a Scania test vehicle. The sensor platform of the vehicle with its illustrative sensor FoV and corresponding reference frames is shown in \Figure\ref{fig:scania-bevda} and \Figure\ref{fig:coordinate-frames} respectively. Each sensor housing contains a lidar with an embedded IMU and two cameras. Note that the cameras were not used in this work.

	Now we describe the necessary notation and reference frames used in our system. The vehicle base frame is denoted as $\Base$ and the world frame is denoted as $\World$. Sensor readings from the lidars and IMUs are represented in their respective sensor frames as $\Lidar^{(k)}$, and $\Imu^{(k)}$ respectively, where, $k \in [\mathtt{F}_\mathtt{L}-\mathtt{Top}, \mathtt{F}_\mathtt{R}-\mathtt{Top}]$ denotes the location of the sensor in the vehicle corresponding to front-left-top and front-right-top respectively. In our discussions, the transformation matrix is denoted as, $\T =
	\left[\begin{array}{ll}\mathbf{R}_{3\times3} & \mathbf{t}_{3\times1} \\
		\mathbf{0}^{\top} & 1\end{array}\right]\in \SEthree$ and
	$\T_{\mathtt{A}\mathtt{B}}$ denotes the transformation matrix of frame $\mathtt{B}$ \textit{wrt} frame $\mathtt{A}$.
	
	\subsection{Problem formulation}
	The primary goal is to estimate the extrinsic calibration between multiple lidar sensors, $\T_{\Lidar^{(k_i)}\Lidar^{(k_j)}}$, in real-time without the need for any calibration targets or odometry estimation. The system is initialized ($\T_{\mathtt{init}}$) only with rough translation component parameters from the vehicle and sensor CAD models. %\mfallon{what does the previous sentence mean. It sounds like `we will initialize with the answer' to me!!!} 
	Finally, the extrinsics are also verified in real-time to ensure robustness.
	%\mfallon{the last sentence doesn't fit in to this conversation... its out of place}

	%\mfallon{dont say `our estimation' ... say `our estimate'. The former is referring to the process/algorithm. the latter to the actual numerical estimate}
	
	\section{Methodology}
	\label{sec:methodology}
	
	Our method starts by matching the noise compensated IMU signals between the base frame, $\Base$, and the IMU frame, $\Imu^{(k)}$, which estimates $\hat{\T}_{\mathtt{IMU}}$. Then a GICP-based \cite{segal2009generalized} point cloud alignment is performed with the initial estimate to compute $\hat{\T}_{\mathtt{GICP}}$, which is further refined by feature matching to estimate $\hat{\T}_{\mathtt{Refined}}$. An additional online verification is performed to ensure the correctness of the calibration process. Thus, $\T_{\Base\Lidar^{(k)}} \simeq \hat{\T}_{\mathtt{Refined}} \times \hat{\T}_{\mathtt{GICP}} \times  \hat{\T}_{\mathtt{IMU}} \times \T_{\mathtt{init}}$. For notional brevity we consider one of the lidars to be base and discuss a general calibration routine between the base frame, $\Base$, and the lidar frame, $\Lidar^{(k)}$.
	
	\subsection{IMU-based initial estimate}
	
	\subsubsection{IMU initialization}	
 	The IMU measurements are in their corresponding sensor frame. For gravity alignment, we use equations eq. 25, 26 from %or initial attitude estimation
 	\cite{pedley2013tilt} to estimate roll and pitch and obtain $\mathtt{R}_{\World\Base}$ after collecting IMU data when the vehicle is static for a few seconds.
	%	\begin{align}
	%		\operatorname{\widehat{Roll}} &=
	%\tan^{-1}\left(\frac{\sum_{i=1}^{t}{\mathbf{f}}_{y_i}}{\sum_{i=1}^{t}\mathbf{f}_{z_i}}\right)
	%\\
	%		\operatorname{\widehat{Pitch}} &=
	%\tan^{-1}\left(\frac{-{\sum_{i=1}^{t}\mathbf{f}}_{x_i}}{\sqrt{\sum_{i=1}^{t}(\mathbf{f}_{y_i}^2
	%+ \mathbf{f}_{z_i}^2)}}\right)
	%	\end{align}
	%where, $\mathbf{f}$ is the linear acceleration from the internally embedded IMU
	%from GNSS system. 

%	Meanwhile, the noise processes and starting bias estimates for the embedded IMU sensor in the lidars and the GNSS unit 
%	were characterized in advance by estimating the Allan Variance 
%	\cite{allan1966statistics} parameters using logs collected while stationary.
%	
%	We compute average of the IMU signals $n$ times in a
%	standstill position for $t$ seconds. The formulation of Allan variance of a
%	signal $\mathbf{s}$, is given as,
%		\begin{align}
%			\operatorname{Allan Variance}(t) &=
%	\frac{1}{2(n-1)}\sum_{i=1}^{n-1}\left[\mathbf{s}(t)_{i+1} -
%	\mathbf{s}(t)_{i}\right]^2 \label{eq:allan-var}\\
%			\operatorname{Allan Deviation}(t) &= \sqrt{\operatorname{Allan Variance}(t)}
%			\label{eq:allan-std}
%		\end{align}
%	
%	 A very interesting read on time-keeping:
%	http://www.allanstime.com/AllanVariance/
%	
%	Allan variance: https://www.cl.cam.ac.uk/techreports/UCAM-CL-TR-696.pdf
	
	\subsubsection{IMU sensor model}
	We have considered a 6-DoF (degree of freedom) IMU such that it has a 3-axis accelerometer and 3-axis gyroscope. The IMU sensor data in its corresponding sensor frame can be represented as,
	\begin{equation}
		\left[\begin{array}{c}\boldsymbol{\omega}_{\Imu_t} \\ \mathbf{a}_{\Imu_t}\end{array}\right]=\left[\begin{array}{c}\boldsymbol{\hat{\boldsymbol{\omega}}_{\Imu_t}} \\ \hat{\mathbf{a}}_{\Imu_t} - \mathbf{R}_{\World\Imu_t}\mathbf{g}_{\World}
		 \end{array}\right]+\left[\begin{array}{l}\mathbf{b}_{\Imu_t}^{\omega} \\ \mathbf{b}_{\Imu_t}^{a}\end{array}\right]+\left[\begin{array}{c}\boldsymbol{\eta}_{\Imu_t}^{\omega} \\ \boldsymbol{\eta}_{\Imu_t}^{a}\end{array}\right]. 
	 \label{eq:imu_model}
	\end{equation}
	Here, $[\boldsymbol{\omega}_{\Imu_t} \in \Real^{3 \times 1}, \mathbf{a}_{\Imu_t} \in \Real^{3 \times 1}]$ represents the measured angular velocity and linear acceleration at timestamp $t$ and $[\hat{\boldsymbol{\omega}}_{\Imu_t}, \hat{\mathbf{a}}_{\Imu_t}]$ represent the latent ideal angular velocity and linear acceleration respectively. $\mathbf{b}_{\Imu_t}^{\omega} \in \Real^{3 \times 1}$ and $\mathbf{b}_{\Imu_t}^{a} \in \Real^{3 \times 1}$ represent the gyro and accelerometer biases which change with time and other factors like temperature. $\boldsymbol{\eta}_{\Imu_t}^{\omega} \sim \mathcal{N}(0, \boldsymbol{\Sigma}_{\omega})$ and $\boldsymbol{\eta}_{\Imu_t}^{a} \sim \mathcal{N}(0, \boldsymbol{\Sigma}_{a})$ are the additive zero-mean white Gaussian noises for gyroscope and accelerometer with covariance $\boldsymbol{\Sigma}_{\omega} \in \Real^{3 \times 3}$ and $\boldsymbol{\Sigma}_{a} \in \Real^{3 \times 3}$ respectively. $\mathbf{g}_{\World} \in \Real^{3 \times 1}$, represents the gravity vector in the world frame, $\World$, and $\mathbf{R}_{\Base\World}$ represent the gravity alignment rotation matrix.
	
	\subsubsection{IMU bias characterization}
	IMU biases are estimated by collecting data sequences at rest. The accelerometer bias is unchanged till the next rest state detection whereas the gyro bias is tracked online with a Kalman filter \cite{kalman1960}. The biases are recomputed whenever a rest state is detected which limits the bias covariance growth.
	%\mfallon{please revise this para as it doesnt make a clear argument IMO} 
	
	\paragraph{IMU state propagation}
	The IMU dynamical model based on angular velocity in quaternion form is shown as,
	\begin{equation}
		\mathbf{q}_{\Base\Imu_t}=\left[\cos \left(\frac{\Delta t}{2}\left\|\boldsymbol{\omega}_{\Imu_t}\right\|\right) \mathbf{I}_{4}+\frac{\sin \left(\frac{\Delta t}{2}\left\|\boldsymbol{\omega}_{\Imu_t}\right\|\right)}{\left\|\boldsymbol{\omega}_{\Imu_t}\right\|}  \boldsymbol{\Omega}_{\Imu_t}\right] \mathbf{q}_{\Base\Imu_{t-1}}
	\end{equation}
	and,
	\begin{equation}
		\boldsymbol{\Omega}_{\Imu_t}=\left[\begin{array}{cccc}0 & {\left[\boldsymbol{\omega}_{\Imu_t}\right]_{z}} & -\left[\boldsymbol{\omega}_{\Imu_t}\right]_{y} & {\left[\boldsymbol{\omega}_{\Imu_t}\right]_{x}} \\ -\left[\boldsymbol{\omega}_{\Imu_t}\right]_{z} & 0 & {\left[\boldsymbol{\omega}_{\Imu_t}\right]_{x}} & {\left[\boldsymbol{\omega}_{\Imu_t}\right]_{y}} \\ {\left[\boldsymbol{\omega}_{\Imu_t}\right]_{y}} & -\left[\boldsymbol{\omega}_{\Imu_t}\right]_{x} & 0 & {\left[\boldsymbol{\omega}_{\Imu_t}\right]_{z}} \\ -\left[\boldsymbol{\omega}_{\Imu_t}\right]_{x} & -\left[\boldsymbol{\omega}_{\Imu_t}\right]_{y} & -\left[\boldsymbol{\omega}_{\Imu_t}\right]_{z} & 0\end{array}\right]. 
	\end{equation}
	$\Delta t$ denotes the sampling period of the IMU data.
	We used the Madgwick filter \cite{madgwick2011} to estimate the refined rotation which minimizes the difference between the measured acceleration and the aligned gravity vector using a gradient descent algorithm as,
	\begin{equation}
		\hat{\mathbf{q}}_{\Base\Imu_t} = \argmin_{\mathbf{q}_{\Base\Imu_t} \in \Real^{4 \times 1}} \left(\begin{array}{l}\tilde{\mathbf{q}}_{\Base\Imu_t} \otimes  {\mathbf{R}_{\Base\world}^{-1}}\mathbf{g}_{\World} \otimes \mathbf{q}_{\Base\Imu_t}- \hat{\mathbf{a}}_{\Imu_t} \\\end{array}\right),
	\end{equation}
	where, $\tilde{\mathbf{q}}$ represents the quaternion conjugate and $\otimes$ is the quaternion product operator. Note that we use the compensated accelerometer signals after bias compensation which is discussed in \ref{sec:imu_bias_est}.
	
	\paragraph{Rest state}
	The rest state is detected if the difference between the norm of the aligned gravity vector and the acceleration vector is less than a predefined threshold, $\tau$ for at least more than 2 seconds. Thus,
	\begin{equation}
	\mathtt{Rest \ detected} = \|\begin{array}{l}\tilde{\hat{\mathbf{q}}}_{\Base\Imu_t} \otimes  \mathbf{g}_{\Base} \otimes \hat{\mathbf{q}}_{\Base\Imu_t}- \hat{\mathbf{a}}_{\Imu_t} \end{array}\|_2 \le \tau	
	\end{equation}
		
	\paragraph{Accelerometer bias estimation}
	\label{sec:imu_bias_est}
	The accelerometer bias is estimated by computing the mean of the signal at rest state and the standard deviation gives us the co-variance of the white Gaussian noise. We recompute these parameters whenever the rest state is detected and keep them unchanged until the next rest state detection. If the rest period duration is $N$ seconds then,
	\begin{equation}
		\begin{aligned}
		\mathbf{b}^{a} &= \frac{\Delta t}{N} \sum_{i=1}^{N}(\mathbf{a}_{\Imu_i} - {\hat{\mathbf{R}}_{\Base\Imu_t}}{\mathbf{R}_{\Base\world}^{-1}}\mathbf{g}_{\World}), \\
		\boldsymbol{\Sigma}_{a} &= \mathtt{diag}\left(\frac{\Delta t}{N}\sum_{i=1}^{N} |\mathbf{a}_{\Imu_i} - \mathbf{b}^{a}|^2\right),\\
		\implies \hat{\mathbf{a}}_{\Imu_t} &= {\mathbf{a}}_{\Imu_t} + {\hat{\mathbf{R}}_{\Base\Imu_t}}{\mathbf{R}_{\Base\world}^{-1}}\mathbf{g}_{\World} - \mathbf{b}^{a} - \mathcal{N}(0, \boldsymbol{\Sigma}_{a}).
		\end{aligned}
	\end{equation}
	
	\paragraph{Gyro bias estimation}
	The initialization of the gyro bias is carried out in a similar fashion as we did for the accelerometer and recomputed whenever the rest state is detected. Thus,
	\begin{equation}
		\begin{aligned}
			\text{System State}, \hat{\mathbf{x}}_0 &= \mathbf{b}^{\omega}_{\Imu_0} = \frac{\Delta t}{N} \sum_{i=1}^{N}(\boldsymbol{\omega}_{\Imu_i})\\
			\text{Measurement Noise}, \mathbf{W}_0 &= \mathcal{N}(0, \boldsymbol{\Sigma}_{\omega}) \\
			\text{Process Noise}, \mathbf{Q} &= (0.05^{\circ}/\mathtt{sec})^2 \eye_{3 \times 3} \\
			\text{Initial Covariance}, \mathbf{P}_0 &= (0.5^{\circ}/\mathtt{sec})^2 \eye_{3 \times 3}.
		\end{aligned}
	\end{equation}
	After that we use a Kalman filter to track the gyroscope bias as a state. The system is modeled by,
	\begin{equation}
		\begin{aligned}
			\mathbf{x}_t & = \hat{\mathbf{x}}_{t-1} + \mathcal{N}(0, \mathbf{Q})\\
			\mathbf{y}_t & = \hat{\mathbf{R}}_{\Base\Imu_t}\mathbf{x}_{t} + \mathbf{W}_t.
		\end{aligned}
	\end{equation}
	The standard Kalman filter update equations become,
	\begin{equation}
		\begin{aligned}
			\mathbf{P}_t & = \mathbf{P}_{t-1} + \mathbf{Q}\\
			\mathbf{K}_t & = \mathbf{P}_t{\hat{\mathbf{R}}_{\Base\Imu_t}} (\boldsymbol{\Sigma}_{\omega} + \hat{\mathbf{R}}_{\Base\Imu_t}\mathbf{P}_t{\hat{\mathbf{R}}_{\Base\Imu_t}^{T}})^{-1} \\
			\hat{\mathbf{x}}_t &= \hat{\mathbf{x}}_{t-1} + \mathbf{K}_t (\mathbf{y}_t - \hat{\mathbf{R}}_{\Base\Imu_t}\hat{\mathbf{x}}_{t-1}) \\
			\mathbf{P}_t & = \mathbf{P}_t - \mathbf{K}_t\hat{\mathbf{R}}_{\Base\Imu_t}\mathbf{P}_t.
		\end{aligned}
	\end{equation}
	Thus after estimating the biases we can rearrange eq. \ref{eq:imu_model} to compute $\hat{\boldsymbol{\omega}}_{\Imu_t}$.
	
	\subsubsection{IMU signal matching}
	\label{sec:imu_calibration}
	
	\paragraph{Rotation estimate}
	%\mfallon{strong correlation is a statistical phrase so isnt proper here. I suggest you mean something like `closely related to'}
	The extrinsics of $\Lidar^{(k)}$ \textit{wrt} $\Base$ frame is closely related to with the extrinsics of $\Imu^{(k)}$ \textit{wrt} $\Base$ frame as they are co-located with known extrinsics from sensor supplier. A rigid body has uniform angular velocity throughout the body. The angular velocities of all the IMU sensors in our sensor arrangement must be equal if their orientations are the same because they are all securely affixed to the vehicle. %\mfallon{the obvious point here is about vibration/flexion across the vehicle. do you want to acknowledge this? - Not relevant to this context imo} 
	Let, [$\hat{\boldsymbol{\omega}}_{\Imu_i^{(k)}}$, $\hat{\mathbf{a}}_{\Imu_i^{(k)}}$] and [$\hat{\boldsymbol{\omega}}_{\Base_i}$, $\hat{\mathbf{a}}_{\Base_i}$] be the estimated angular velocity and linear acceleration of the $k^{th}$ IMU and the base IMU at timestamp $t_i$ after bias compensation. By removing the $k^{th}$ superscript for brevity, our optimization problem becomes,
	\begin{equation}
		\begin{aligned}
			\R_{\Base\Imu}^{\star} &= \argmin_{\R_{\Base\imu}} \sum_{i=1}^{N}
			\| \R_{\Base\imu} \hat{\boldsymbol{\omega}}_{\Base_i} - \hat{\boldsymbol{\omega}}_{\Imu_i} \|^2_{\boldsymbol{\sum_i}}, \\
			& \textit{s.t.} \ \ \R_{\Base\Imu} \R_{\Base\Imu}^T = \Identity_3,
			\label{eq:imu_R_calibration}
		\end{aligned}
	\end{equation}
	which, can be simply solved with Kabsch alignment \cite{Kabsch1978}.
	
	%Mean Center Data
	%Ac = mean(A')';
	%Bc = mean(B')';
	%A = A-repmat(Ac,1,n);
	%B = B-repmat(Bc,1,n);
	
	%Calculate Optimal Rotation
	%[u,~,v]=svd(A*B');
	%R = v*\text{diag}([1 1 \text{det}(v*u')])*u';

\begin{figure}[!h]
		\centering
		\includegraphics[width=1\columnwidth]{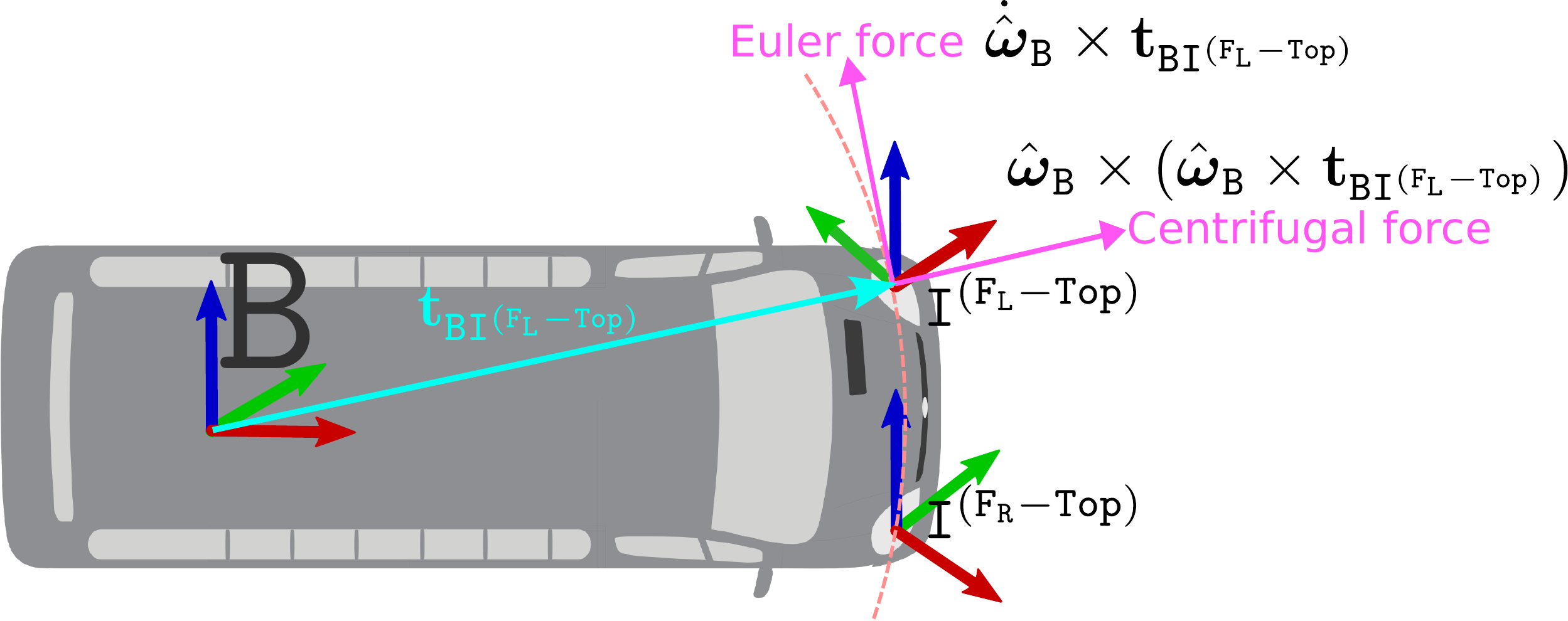}
		%\vspace{-5mm}
		\caption{Illustration of IMU signal transformation between the base frame, $\Base$, and the IMU frame, $\Imu^{(\mathtt{F}_\Lidar-\mathtt{Top})}$.} 
		\label{fig:imu-rig}
		%\vspace{-5mm}
	\end{figure}

	\paragraph{Translation estimate}
	For the acceleration components we account for the Coriolis forces as illustrated in \Figure\ref{fig:imu-rig} and equate the translation components as,
	\begin{equation}
		\begin{aligned}
			{(\mathbf{R}_{\Base\Imu}^{\star})}^{-1}\hat{\mathbf{a}}_{\Imu} &=\hat{\mathbf{a}}_\Base+\underbrace{\hat{\boldsymbol{\omega}}_\Base
			\times\left(\hat{\boldsymbol{\omega}}_\Base \times \mathbf{t}_{\Base\Imu}\right)}_{\text
			{Centrifugal force }}+\underbrace{\dot{\hat{\boldsymbol{\omega}}}_\Base \times
			\mathbf{t}_{\Base\Imu}}_{\text {Euler force }} \\
			&= \hat{\mathbf{a}}_\Base+[\hat{\boldsymbol{\omega}}_\Base]^2_{\times} \mathbf{t}_{\Base\Imu} +
			[\dot{\hat{\boldsymbol{\omega}}}_\Base]_{\times}\mathbf{t}_{\Base\Imu},
			\label{eq:coriolis}
		\end{aligned}
	\end{equation}
	where, $[\boldsymbol{.}]_{\times}$ is a skew-symmetric matrix and $[\boldsymbol{a}]_{\times} \boldsymbol{b} = \boldsymbol{a} \times \boldsymbol{b}$.
	
	\noindent Since, we have already computed $\mathbf{R}^\star_{\Base\Imu}$, our optimization problem for the translation component becomes,
	\begin{align}
		\tran_{\Base\Imu}^{\star} = \argmin_{\tran_{\Base\imu}}\sum_{i=1}^{N}
		\resizebox{0.68\columnwidth}{!}{$
			\| \underbrace{(\left[\boldsymbol{\hat{\omega}_{\Base_i}}\right]^2_{\times} + [\dot{\hat{\boldsymbol{\omega}}}_{\Base_i}]_{\times})}_{\mathbf{A}_i} \mathbf{t}_{\Base\Imu} - \underbrace{(\R^{-1\star}_{\Base\imu} \hat{\boldsymbol{a}}_{\imu_i} - \hat{\boldsymbol{a}}_{\Base_i})}_{\mathbf{B}_i}  \|^2,
			\label{eq:imu_t_calibration}$}
	\end{align}
	where, $\mathbf{A} = [\mathbf{A}_0 \ \mathbf{A}_1 \ \dots \ \mathbf{A}_n]^T$, $\mathbf{B} = [\mathbf{B}_0 \ \mathbf{B}_1 \ \dots \ \mathbf{B}_n]^T$ and $\mathbf{x} = \mathbf{t}_{\Base\Imu}$ $\forall i \in [1, N]$. This is a system of linear equations of the form, $\mathbf{A}{\mathbf{x}} = \mathbf{B}$, which can be solved using a least-square approach as,
	\begin{equation}
		\mathbf{x}^\star = \argmin_{\mathbf{x}}\| \mathbf{A}\mathbf{x} - \mathbf{B}\|^2_{\boldsymbol{\Sigma}},
		\label{eq:unbounded}
	\end{equation}
	where, ${\boldsymbol{\Sigma}}$ denotes the co-variance of the residual. The main approach used to tackle these problems \cite{nocedal1999numerical} involves solving a series of approximations to the original problem repeatedly by linearizing as, $F(\mathbf{x}+\Delta \mathbf{x}) \approx F(\mathbf{x})+\mathbf{J}(\mathbf{x}) \Delta \mathbf{x}$, where, $J$ being the jacobian of $F(\mathbf{x})$. In this way $\hat{\mathbf{x}}$ is updated in the current iteration as,
	$\hat{\mathbf{x}} \leftarrow \hat{\mathbf{x}} \boxplus \mathbf{\Delta x}$, where $\boxplus$ is an addition operator in the manifold and the problem becomes,
	\begin{equation}
		\begin{aligned}
			&\mathbf{\Delta x}^\star = \argmin_{\mathbf{\Delta x}}\frac{1}{2}\| (\mathbf{A}\hat{\mathbf{x}} - \mathbf{B}) + \mathbf{A}\mathbf{\Delta x}\|^2_{\boldsymbol{\Sigma}}.
			\label{eq:unbounded_expanded}
		\end{aligned}
	\end{equation}
	The optimal solution is given by,
	\begin{equation}
		\underbrace{\left( \mathbf{J}^{\top} \boldsymbol{\Sigma}^{-1}\mathbf{J} \right)}_{\text{Fisher information matrix}}\mathbf{\Delta x}=-\mathbf{J}^{\top} \boldsymbol{\Sigma}^{-1} (\mathbf{A}\hat{\mathbf{x}} - \mathbf{B}).
		\label{eq:bvls_iterative_solution}
	\end{equation}
	%We used a bounded variable least-square solver \cite{Agarwal_Ceres_Solver_2022} with trust region approach and updated the parameters only if it improves the cost function, as defined in \eq~\ref{eq:error_term}.
	In practice, since we can take a reliable initial estimate for the translation component from the CAD parameters (within $\approx$10 cm), %(\mfallon{give a degree of accuracy: within Xcm}), 
	we search for $\mathbf{x}^\star$ only in a local neighborhood within reasonable bounds. Thus, we solve a bounded variable least squares problem as,
	\begin{equation}
		\mathbf{x}^\star = \argmin_{\mathbf{L} \le \mathbf{x} \le \mathbf{U}}\| \mathbf{A}\mathbf{x} - \mathbf{B}\|^2_{\boldsymbol{\Sigma}},
		\label{eq:bvls}
	\end{equation}
	where, $\mathbf{L}$ and $\mathbf{U}$ are the lower and upper bounds of $\mathbf{x}$ respectively. Thus the solution space in \eq~\ref{eq:unbounded_expanded} is modified with an additional constraint as, 
	\begin{equation}
		\begin{aligned}
			&\mathbf{\Delta x}^\star = \argmin_{\mathbf{\Delta x}}\frac{1}{2}\| (\mathbf{A}\hat{\mathbf{x}} - \mathbf{B}) + \mathbf{A}\mathbf{\Delta x}\|^2_{\boldsymbol{\Sigma}},\\
			& \textit{s.t.} \ \mathbf{L} \le \hat{\mathbf{x}} + \mathbf{\Delta x} \le \mathbf{U}.
			\label{eq:bvls_expanded}
		\end{aligned}
	\end{equation} 
	
	\paragraph{Observability analysis}
	Not all vehicle motions are suitable for exciting the degrees of freedom to allow calibration. As a result, it's important to identify suitable motion segments for information-aware calibration updates. We captured this information by comparing the angular velocity between $\Imu^{(k)}$ and $\Base$ frame based on \eq\ref{eq:imu_R_calibration} which can also be solved iteratively with the update step as,
	\begin{equation}
		\resizebox{\columnwidth}{!}{$
		\underbrace{\left(\sum_{i} \mathbf{J}_{i}^{\top} \Sigma_{i}^{-1} \mathbf{J}_{i}\right)}_{\text{Fisher information matrix}} \mathbf{\Delta}{\mathbf{\R}}_{\Base\imu} \stackrel{\eq~\ref{eq:bvls_iterative_solution}} {=} -\sum_{i} \mathbf{J}_{i}^{\top} \Sigma_{i}^{-1} (\hat{\mathbf{\R}}_{\Base\imu} \boldsymbol{\omega}_{\Base_i} - \boldsymbol{\omega}_{\Imu_i})$}
		\label{eq:residual_calc}
	\end{equation}
	where, ${\boldsymbol{\Sigma}_i} = \operatorname{cov}(\hat{\mathbf{\R}}_{\Base\imu} \boldsymbol{\omega}_{\Base_i} - \boldsymbol{\omega}_{\Imu_i})$ and $\mathbf{J}_{i} = \boldsymbol{\omega}_{\Base_i}^T$. The Fisher information matrix, $\mathcal{I}_{N\times N}$ captures all the information contained in the measurements. We do the processing after arranging the data in a batch size of $N$. We perform a Singular Value Decomposition of $\mathcal{I}_{N\times N}$ for each batch as:
	\begin{equation}
		\mathcal{I}_{N\times N} = \textbf{USU}^T,
	\end{equation}
	where, $\textbf{U} = [\textbf{u}_1, \textbf{u}_2 \dots , \textbf{u}_N]$ and $\textbf{S} = \text{diag}(\sigma_1, \sigma_2 \dots , \sigma_N)$ is a diagonal matrix of singular values in decreasing order. Information about the data in the batch is indicated by the value of the minimal singular value. If the minimal singular value exceeds a certain threshold (design decision), we may say that there is sufficient excitation %\mfallon{this should be singular - as its a continuous value} 
	in the batch of data to allow for extrinsics computation and hence chosen for calibration.

	\subsection{Point cloud based alignment}
	
	\paragraph{Initial match}
	%\mfallon{you have changed tense from present to past here. "We capture" before; now "we filtered"}
	Based on our sensor setup we first filtered the point clouds with a box filter to retain the region comprising of maximum overlap between the two point clouds for faster processing. For our setup, we filtered the point clouds between $[0, -10, -10]m$ and $[50, 10, 10]m$. Then the initial guess from the IMU-based matching was used as a prior for the GICP-based \cite{segal2009generalized} matching on the filtered point clouds. 
	
	\paragraph{Line and plane feature extraction and matching}
	After obtaining improved extrinsic parameters from GICP we further refined the extrinsics using line feature extraction and matching. First, segmentation \cite{shan-legoloam} and local curvature \cite{Bogoslavskyi2016} of each point in the filtered clouds are evaluated. The points with the highest and lowest curvature are allocated to the sets of line and plane clouds, respectively. Then, we fit the line and plane model to the corresponding line and plane cloud sets using PROSAC \cite{chum2005matching} which exploits the spatial coherence of the line and plane points. Thus, for each lidar point cloud, we obtained the line and plane feature set as $\calF^L$ and $\calF^P$, where a line and plane are modeled as,
	
	\begin{equation}
		\begin{aligned}
		\calF^L_{\Lidar^{(k)}} &:= \left\{\langle \hat{\mathbf{n}}^{(l)}_{\Lidar^{(k)}},\hat{\mathbf{c}}^{(l)}_{\Lidar^{(k)}} = (l_x, l_y, l_z)\rangle \in \SOthree \times
		\mathbb{R}^{3}\right\} \\
		\calF^P_{\Lidar^{(k)}} &:=  \left\{\langle \hat{\mathbf{n}}^{(p)}_{\Lidar^{(k)}},\hat{\mathbf{c}}^{(p)}_{\Lidar^{(k)}} = (p_x, p_y, p_z)\rangle \in \SOthree \times
		\mathbb{R}^{3}\right\}
		\end{aligned}
	\end{equation}
	where, $\hat{\mathbf{n}}$ and $\hat{\mathbf{c}}$ denotes the estimated direction and center of the line and plane respectively. Finally, additional filtering is performed where only the closest line and planes between the set of different point clouds are kept for further refinement and verification. Two lines or planes are considered a close match if,
	\begin{equation}
	\begin{aligned}
		\alpha &= \|\cos^{-1}(\hat{\mathbf{n}}_{\Base} \cdot \hat{\mathbf{n}}_{\Lidar^{(k)}})\| &\le \tau_\alpha\\ 
	\delta &= \frac{\left|(\hat{\mathbf{n}}_{\Base} \times \hat{\mathbf{n}}_{\Lidar^{(k)}} )\cdot\left(\hat{\mathbf{c}}_{\Base}-\hat{\mathbf{c}}_{\Lidar^{(k)}}\right)\right|}{\| (\hat{\mathbf{n}}_{\Base} \times \hat{\mathbf{n}}_{\Lidar^{(k)}} ) \|} & \le \tau_\delta\\
	\end{aligned}
	\end{equation}

    % Testing with graph-cut ransac as well, but in progress. Add if finished.
	% https://github.com/danini/graph-cut-ransac/
	% Opencv (4.x branch):  https://github.com/opencv/opencv/blob/c9a4775d49202538ffa40fc147b502049a2ae93c/modules/calib3d/src/usac/ransac_solvers.cpp#L1104C14-L1104C27 
	
	\paragraph{Extrinsic refinement}
	\label{sec:refinement}
	For refinement, we chose $N$ set of corresponding line features from the point clouds for which $\tau_\alpha^{(l)} = 5^\circ$ and $\tau_\delta^{(l)} = 0.5m$. By removing the $k$ superscript for brevity, the optimal rotation can be computed by solving the following optimization,  
	\begin{align}
		\hat{\mathbf{R}}_{\mathtt{Refined}} &=
		\argmin_{\mathbf{R}_{\Base\Lidar}}\sum_{i=1}^{N}
		\|\mathbf{R}_{\Base\Lidar}\hat{\mathbf{n}}_{\Base_i} - \hat{\mathbf{n}}_{\Lidar_i}\|^2_F
		\label{eq:rot_cost_function}\\
		&= \argmin_{\mathbf{R}_{\Base\Lidar}}\sum_{i=1}^{N}2\left[ 1 -
		\operatorname{Tr}\left(\mathbf{R}_{\Base\Lidar}\hat{\mathbf{n}}_{\Base_i}\hat{\mathbf{n}}_{\Lidar_i}^T\right)\right].
		\label{eq:optimal_rotation}
	\end{align}
	We formulate our problem as a Q-method and convert the cost function based on rotation matrices to quaternion form as discussed in our earlier work \cite{das2023observability} as,
	\begin{align}
		%&\stackrel{eq. \ref{eq:optimal_rotation_cost_function}} {=}
		%-\sum_{i=1}^{N}\operatorname{Tr}\left(\mathbf{R}_{\Base\Lidar}\mathbf{R}_{\Base_i}\mathbf{R}_{\Lidar_i}^T\right)
		%\nonumber \\
		%&=
		%-\operatorname{Tr}\left[\sum_{i=1}^{N}\left(\mathbf{R}_{\Base\Lidar}\mathbf{R}_{\Base_i}\mathbf{R}_{\Lidar_i}^T\right)\right]\nonumber
		%\\
		\resizebox{0.87\columnwidth}{!}{$J(\mathbf{R}_{\Base\Lidar})
			\stackrel{\eq~\ref{eq:optimal_rotation}} {=}
			-\operatorname{Tr}\left[\mathbf{R}_{\Base\Lidar}
			\sum_{i=1}^{N}\left(\hat{\mathbf{n}}_{\Base_i}\hat{\mathbf{n}}_{\Lidar_i}^T\right)\right] = -\operatorname{Tr}\left[\mathbf{R}_{\Base\Lidar}
			\mathbf{\Delta}_{3\times3}\right]
			\label{eq:rotation_cost_function}$}.
	\end{align}
	The attitude quaternion which minimizes the cost function is the unit
	eigenvector corresponding to the largest eigenvalue of $\mathbf{K}$ (eigenvalues of a symmetric matrix are real and hence can be sorted). Thus,
	\begin{align}
		\hat{\mathbf{q}}(\mathbf{R}_{\Base\Lidar}) &=
		\argmax_{\operatorname{Eigenvalues}(\mathbf{K})}{\operatorname{Eigenvectors}(\mathbf{K})}
		\label{eq:R_estimated},
	\end{align}
	where, $\mathbf{K} =
	\left[\begin{array}{cc}\mathbf{\Gamma}-\mu \mathbf{I} & \mathbf{\Lambda} \\
		\mathbf{\Lambda}^{T} & \mu\end{array}\right]$, $\mathbf{\Lambda}=\left[\begin{array}{ccc} \mathbf{\Delta}(2,3)
		-\mathbf{\Delta}(3, 2) \\ \mathbf{\Delta}(3, 1) - \mathbf{\Delta}(1, 3) \\
		\mathbf{\Delta}(1, 2)-\mathbf{\Delta}(2, 1)\end{array}\right]$,  $\mathbf{\Gamma} = \mathbf{\Delta}^T+\mathbf{\Delta}$ and $\mu =
	\operatorname{Tr}(\mathbf{\Delta})$.
	
	The optimal translation is obtained by matching the line centers as,
	\begin{align}
		\hat{\mathbf{t}}_{\mathtt{Refined}} = \frac{1}{N}\sum_{i=1}^{N}
		\left(\hat{\mathbf{c}}_{\Lidar} - \hat{\mathbf{c}}_{\Base}\right).
		\label{eq:trans_cost_function}
	\end{align}	
	
	\paragraph{Extrinsics verification}
	After estimating the refined extrinsics we obtained the final extrinsic parameters as $\T_{\Base\Lidar^{(k)}} \simeq \hat{\T}_{\mathtt{Refined}} \times \hat{\T}_{\mathtt{GICP}} \times  \hat{\T}_{\mathtt{IMU}} \times \T_{\mathtt{init}}$. At this stage, we applied the transformation to the point clouds to orient them to a common reference frame and extract plane features. If for the corresponding plane features, $\tau_\alpha^{(p)} = 1^\circ$ and $\tau_\delta^{(p)} = 0.3m$, we accept the final extrinsic calibration, otherwise, we recompute the whole point-based alignment.
	
	\section{Experimental Results}
    We carried out a series of experiments to test different aspects of algorithm --- starting from a basic sensor rig and building up to a full demonstration on a test vehicle.

	We first carried out MIMU calibration by mounting 2 IMUs on a custom made planar board. After this we collected sensor data from the test vehicle and computed the baseline extrinsics, where the (lidar) odometry was estimated using Fast-LIO2 \cite{xu2022fast} and matched using Kabsch alignment \cite{Kabsch1978}.

	We then benchmark our results against CROON \cite{wei2022croon} where ground plane extraction and ICP (with normal) was performed between corresponding lidar frames and averaged our results across all the frames in the scene. Finally, we performed our calibration experiments with IMU-based initialization. The results are presented as the difference between ground truth (GT) and estimated extrinsics. 

	\subsection{MIMU calibration on a test rig}
\begin{figure}[!hbt]
		\vspace{-0.2cm}
		\centering
		\begin{multicols}{2}
			\includegraphics[width=1.0\linewidth,trim=10 0 60 0, clip]{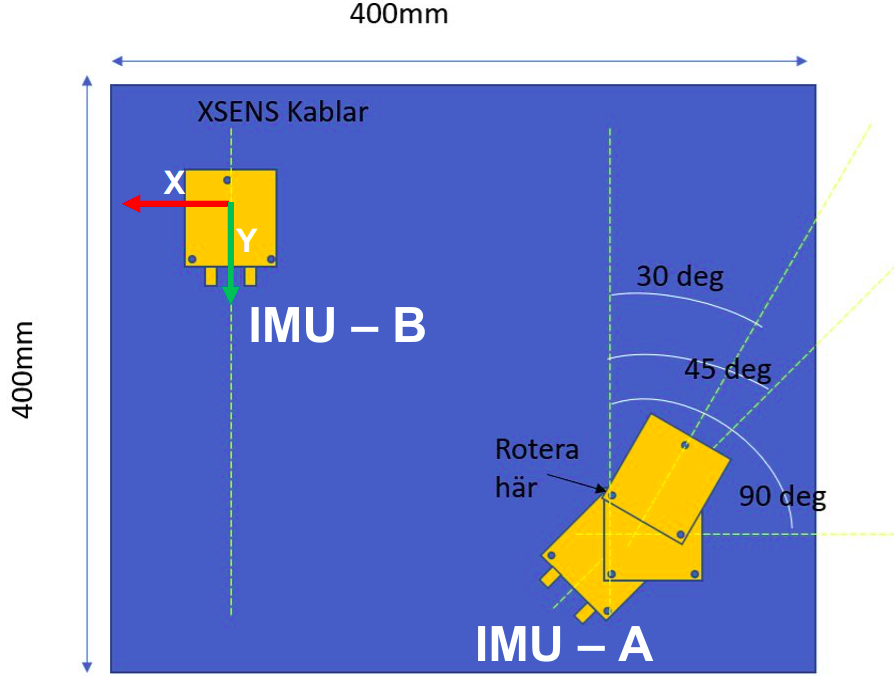}
			\subcaption{IMU-B is kept fixed whereas, IMU-A is rotated clockwise at $30^{\circ}, 45^{\circ}$ and $90^{\circ}$ configurations.}
			\label{fig:imu_board_convcept}
			\includegraphics[width=1.0\linewidth,scale=1.0]{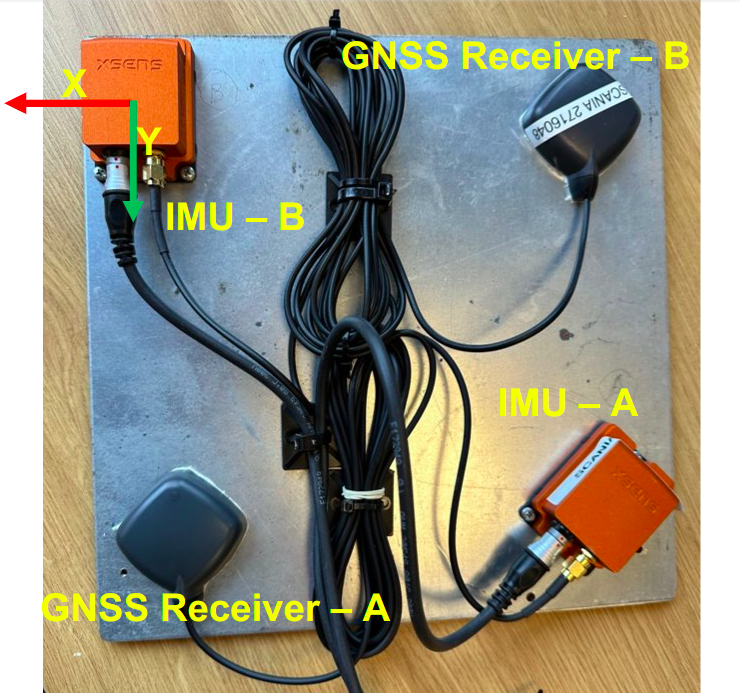}
			\subcaption{$45^{\circ}$ configuration example.}
			\label{fig:imu_board_45}
		\end{multicols}
	\vspace{-0.3cm}
	\caption{Illustration of the test IMU board.}
	\label{fig:imu_board}
	\vspace{-0.2cm}
	\end{figure}

\begin{figure}[!h]
		\centering
		\includegraphics[width=1\columnwidth]{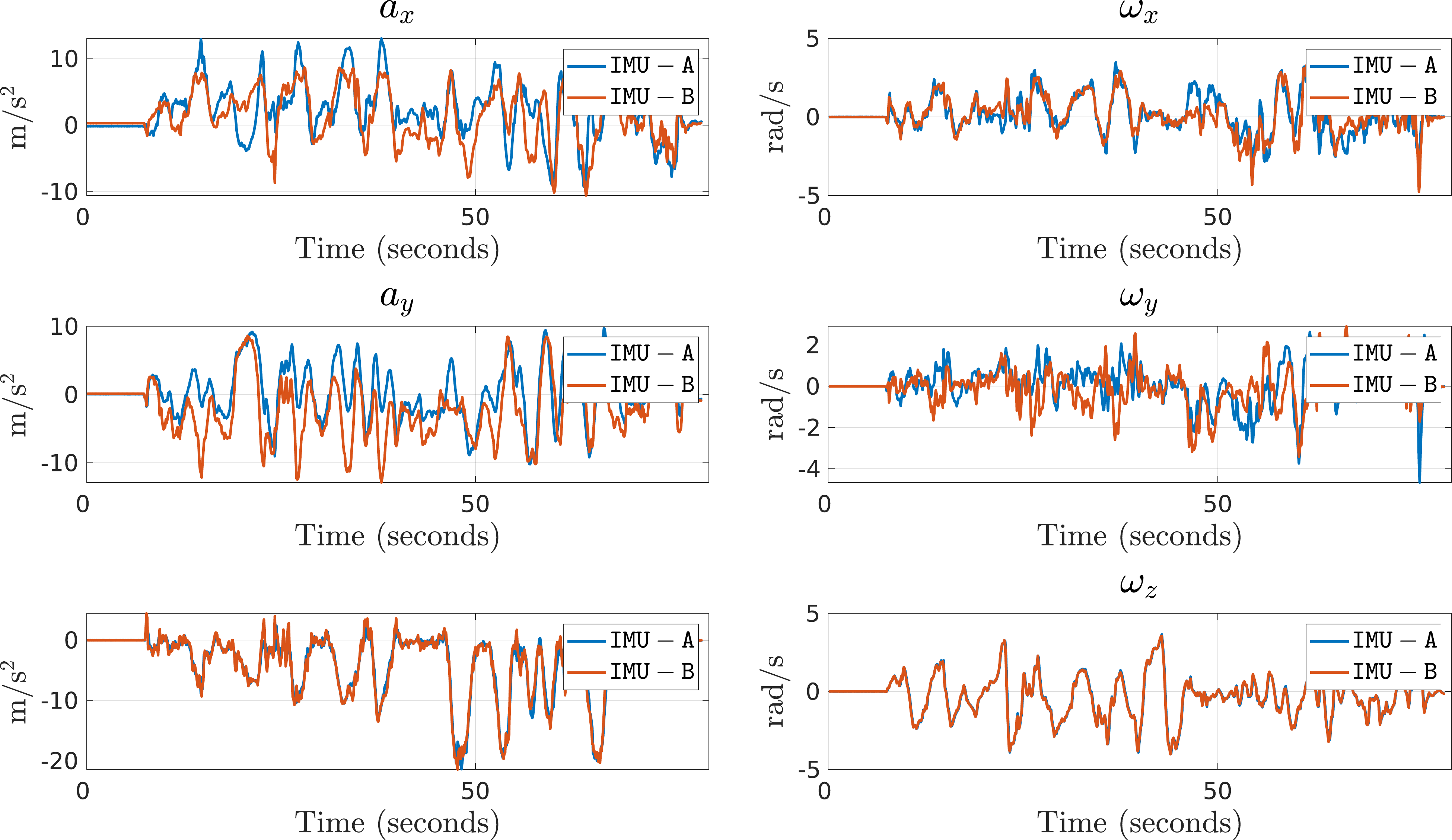}
		%\vspace{-5mm}
		\caption{Raw IMU signals from $\mathtt{IMU-A}$ and $\mathtt{IMU-B}$ at $45^{\circ}$ configuration in individual sensor reference frame.}
		\label{fig:raw_signal}
		\vspace{-4mm}
	\end{figure}

\begin{figure}[!h]
		\centering
		\includegraphics[width=1\columnwidth]{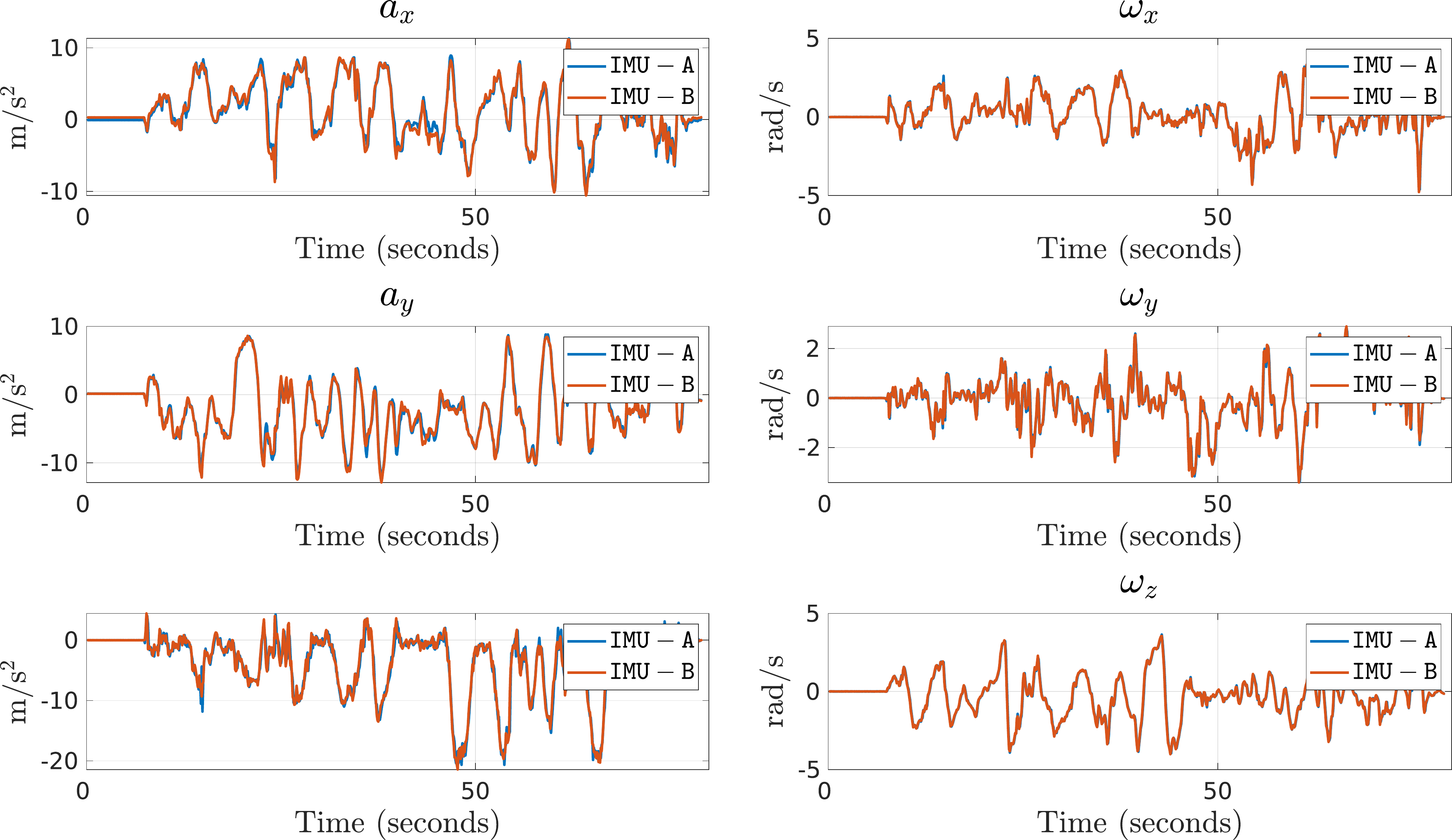}
		%\vspace{-5mm}
		\caption{Calibrated IMU signals from $\mathtt{IMU-A}$ and $\mathtt{IMU-B}$ at $45^{\circ}$ configuration with bias compensation in $\mathtt{IMU-B}$ frame.}
		\label{fig:bias_compensated_signal}
		\vspace{-4mm}
	\end{figure}

	We created the planar test rig seen in \Figure\ref{fig:imu_board} to verify the MIMU calibration technique. The board contained 2 Xsens MTi-300 IMUs along with GNSS receivers to ensure time synchronization. $\mathtt{IMU-B}$ was fixed to the board while we could adjust the pose of $\mathtt{IMU-A}$ clockwise to three different orientations --- $30^{\circ}, 45^{\circ}$ and $90^{\circ}$. We collected IMU data at each orientation and computed the extrinsics by matching the IMU signals as discussed in \ref{sec:imu_calibration}. We measured the GT translation components between the IMU centers using a measuring tape. %The sample board, although planar, had some deviations in pitch and roll alignment that we were unable to measure. \mfallon{the previous sentence isnt good - how hard is it to make a metal plate flat! Can you remove this comment?} 
	The results are shown in Table~\ref{tab:calib_results_imu_dataset} and an illustration of the signals before and after calibration of the $45^\circ$ configuration are shown in \Figure\ref{fig:raw_signal} and \Figure\ref{fig:bias_compensated_signal} respectively. We see that after the calibration steps the IMU signals show good qualitative alignment when plotted in the same reference frame.
	%\mfallon{You never refer to the plots of fig 5 and 6. You need to at least comment on these results and say the plots show good qualitative alignment.}
	
	\normalsize
	\begin{table}[!h]
		\centering
		\resizebox{0.99\columnwidth}{!}{
			\begin{tabular}{c|cccccc}  \toprule
				\multicolumn{7}{c}{Performance comparison at different configurations: IMU test board dataset}\\
				\midrule
				Configuration & $\Delta x [m]$  & $\Delta y [m]$ & $\Delta z [m]$ & $\Delta roll [^{\circ}]$ & $\Delta pitch [^{\circ}]$ & $\Delta yaw [^{\circ}]$\\
				\midrule
				$30^{\circ}$ & -0.0500 & -0.0461 & 0.0015 & -0.3268 & -2.3129 & -0.8808\\
				\midrule
				$45^{\circ}$ & -0.0950 & 0.1018 & 0.0018 & -0.4305 & -2.2219 & -1.2211\\
				\midrule
				$90^{\circ}$ & -0.0003 & 0.0146 & -0.0065 & -0.3174 & -2.3144 & -0.0562 \\
				\bottomrule
		\end{tabular}}
		\caption{}
		\label{tab:calib_results_imu_dataset}
		\vspace{-8mm}
	\end{table}

	%\mfallon{having these 12 plots doesnt seem to achieve much IMO.}

	\subsection{Lidar extrinsic calibration using IMU-based initialization on Scania datasets}
	\label{sec:dataset}
    %\mfallon{you need to explain what the point of this experiment was before starting!}
	After verifying the basic MIMU calibration approach on the test rig, we used the same method to pre-initialize our lidar extrinsic calibration algorithm. The GT for the extrinsic calibration parameters was pre-computed offline. We analyzed our lidar calibration results on 4 different test sequences which are described in Table~\ref{tab:dataset}, covering different motion scenarios. We consider the feature richness level to be low if there are very few ($\le$ 3) geometric features detected per frame on average. In Seq-1, we performed aggressive maneuvers to excite all possible degrees of freedom in a parking lot. Seq-2 consists of very slow driving ($\approx$\SI{10}{\kilo\meter\per\hour}) in a feature rich environment.

	Seq-3 and Seq-4 contain data from an urban driving scenario ($\approx$\SI{40}{\kilo\meter\per\hour}) in normal traffic conditions.	We also computed the lidar odometry using FAST-LIO2 \cite{xu2022fast}, for all the sequences and computed the RMSE of the absolute pose error (APE) between the lidars using GT extrinsics. This helped us to create the baseline for the extrinsics computation using Kabsch alignment. The high APE supports our claim that because the pose alignment technique has inherent drift and computationally heavy, it does not bring any additional value to the target-less extrinsics estimation process other than just a rough initialization.

	%\mfallon{is it Kabsch or Kabsh? you use both in the paper}

	\begin{table}[!h]
		\centering
		%\vspace{2mm}
		%\fontsize{18}{18}\selectfont
		\resizebox{\columnwidth}{!}{
			\begin{tabular}{l|ccccc}  \toprule
				%\begin{tabular*}{\textwidth}{@{\extracolsep{\fill}}l|cccc}\toprule
				\multicolumn{6}{c}{Scania dataset collection details for the experiments}\\
				\midrule \midrule
				\textbf{Data} & Scenario & Length (\si{Km}) & Duration (\si{secs}) & Feature richness & APE\\
				\midrule
				Seq-1 & 8-pattern in parking & 0.325 & 72.5 & Low & 2.390 \\
				Seq-2 & Slow turning & 0.333 & 128.7 & Medium & 2.152 \\
				Seq-3 & Urban driving & 0.439 & 78.1 & High & 1.886 \\
				Seq-4 & Urban driving & 2.209 & 281.4 & High & 5.793 \\
				\bottomrule
		\end{tabular}}
		%\vspace{-3mm}
		\caption{}
		\label{tab:dataset}
		\vspace{-4mm}
	\end{table}

	\subsection{Bias estimation results}
	To improve our calibration performance, which relies on the signal-to-signal matching policy, we compensate for the biases in both the accelerometer and angular velocity signals. When a period of rest is identified, the accelerometer biases are recomputed and held constant for the following period. The angular velocity biases are updated based on the estimated orientation using the Madgwick filter that uses the bias-compensated accelerometer signals for gravity alignment. Whenever a period of rest is detected the covariances of the angular velocity biases converge.
		
	\begin{figure}[!h]
		\centering
		\includegraphics[width=1\columnwidth]{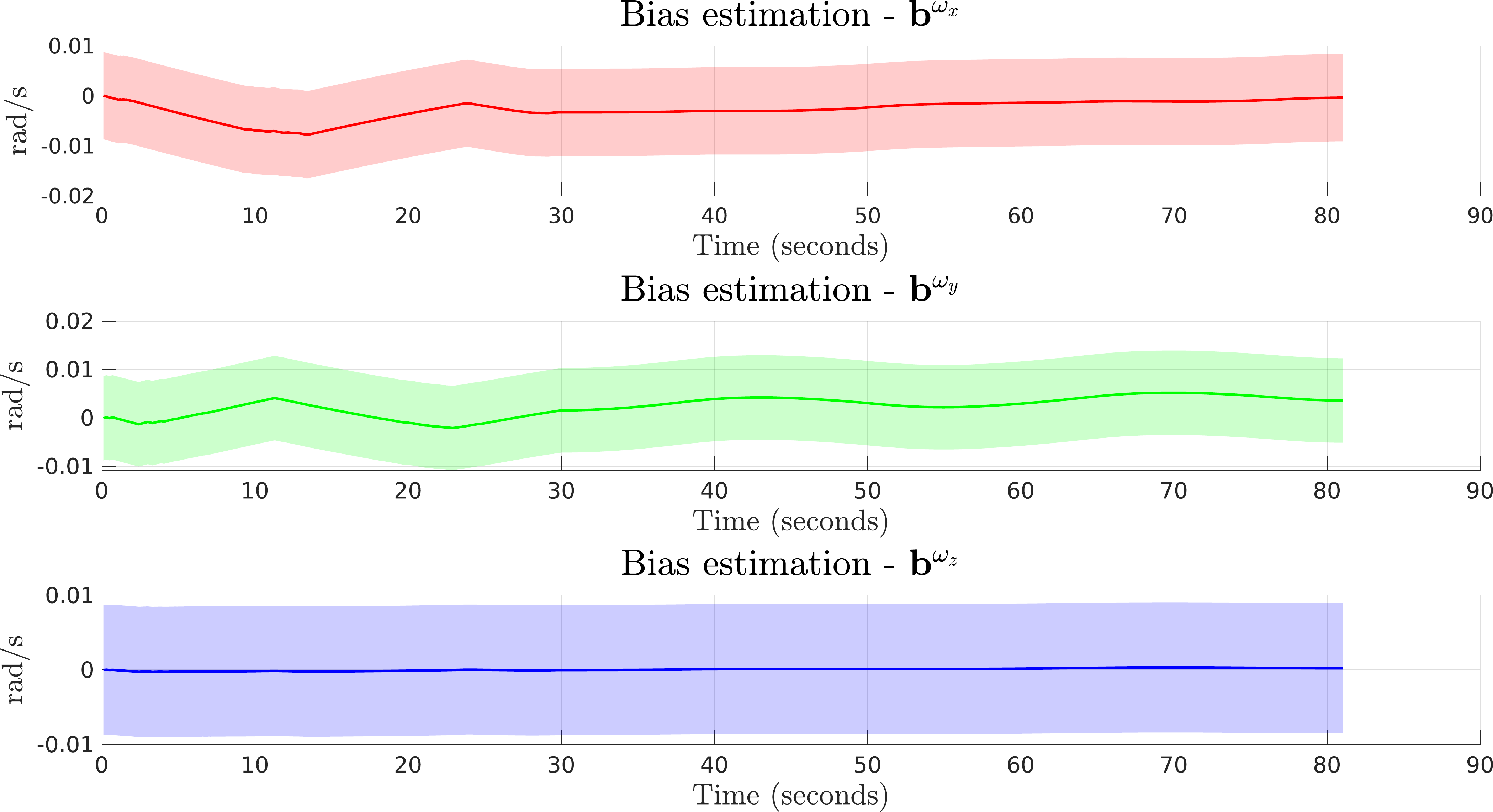}
		%\vspace{-5mm}
		\caption{Bias estimation for the $\mathtt{F}_\mathtt{L}-\mathtt{Top}$ angular velocity signals for Seq-1.}
		\label{fig:bias_estimation_1}
		\vspace{-4mm}
	\end{figure}
	
	\begin{figure}[!h]
		\centering
		\includegraphics[width=1\columnwidth]{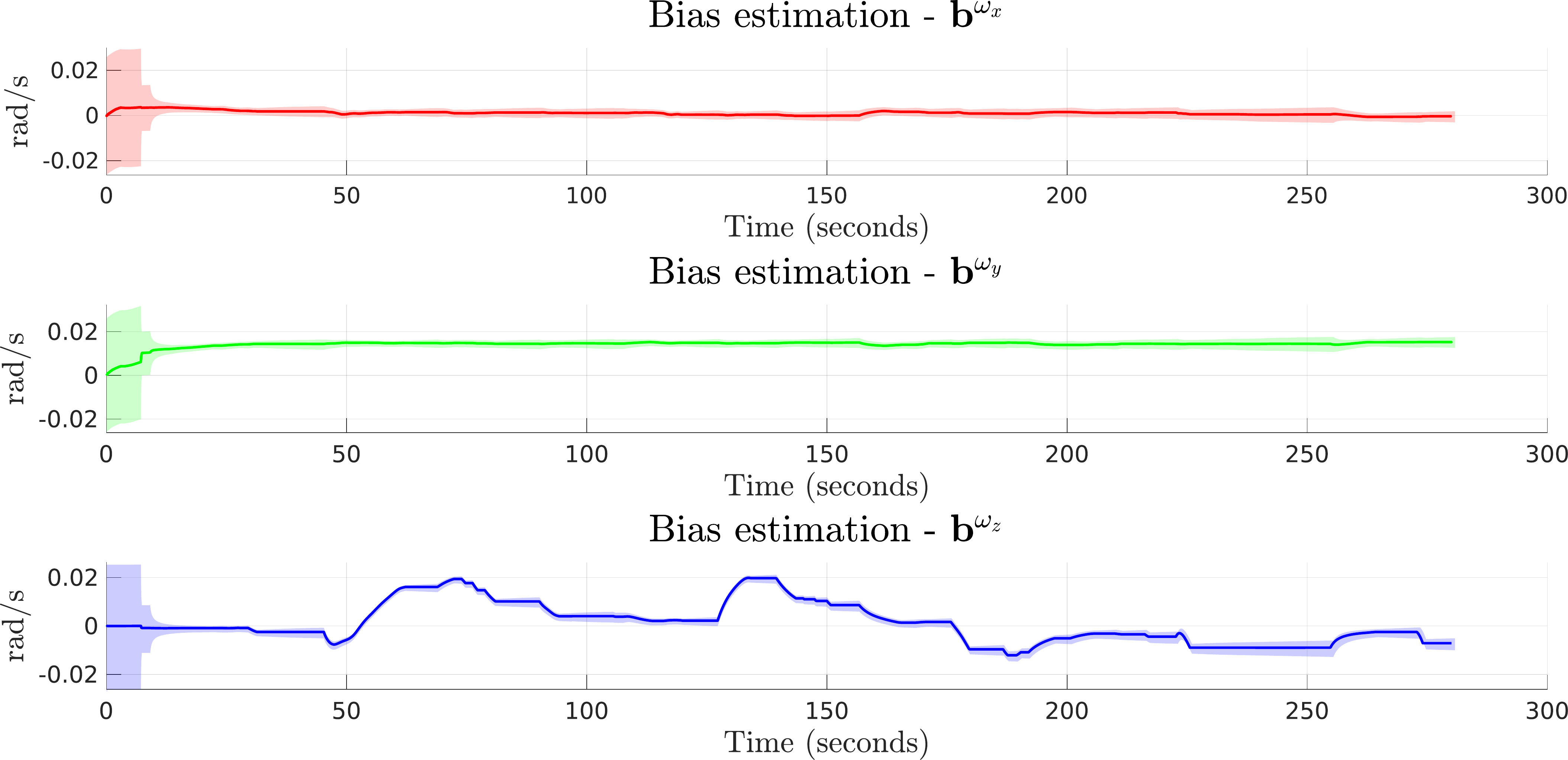}
		%\vspace{-5mm}
		\caption{Bias estimation for the $\mathtt{F}_\mathtt{L}-\mathtt{Top}$ angular velocity signals for Seq-4.}
		\label{fig:bias_estimation_4}
		%\vspace{-4mm}
	\end{figure}
	In Seq-1, we didn't have any periods of rest. Because of this, the angular velocity bias covariances did not converge as seen in \Figure\ref{fig:bias_estimation_1}. Whereas, in \Figure\ref{fig:bias_estimation_4}, we see that the angular velocity bias covariances converge after the detection of the rest period for a few seconds at the beginning of Seq-4.
	
	\subsection{Observability analysis}	
	To study observability analysis, we extracted the information matrix by comparing the angular velocity signals between the IMU, $\Imu^{(k)}$, and the $\Base$ frame by dividing the data into equal segments of 10 sec each. We analyze the SVD of the Fisher information matrix from \eq\ref{eq:residual_calc} and select the IMU data in the segment for calibration if the minimum singular value is
	greater than a threshold. For our experimental dataset, we set the threshold as $5^{-10}$.
	
	\begin{figure}[!h]
		\centering
		\begin{subfigure}[b]{1\columnwidth}
			\centering
			\includegraphics[width=1\columnwidth]{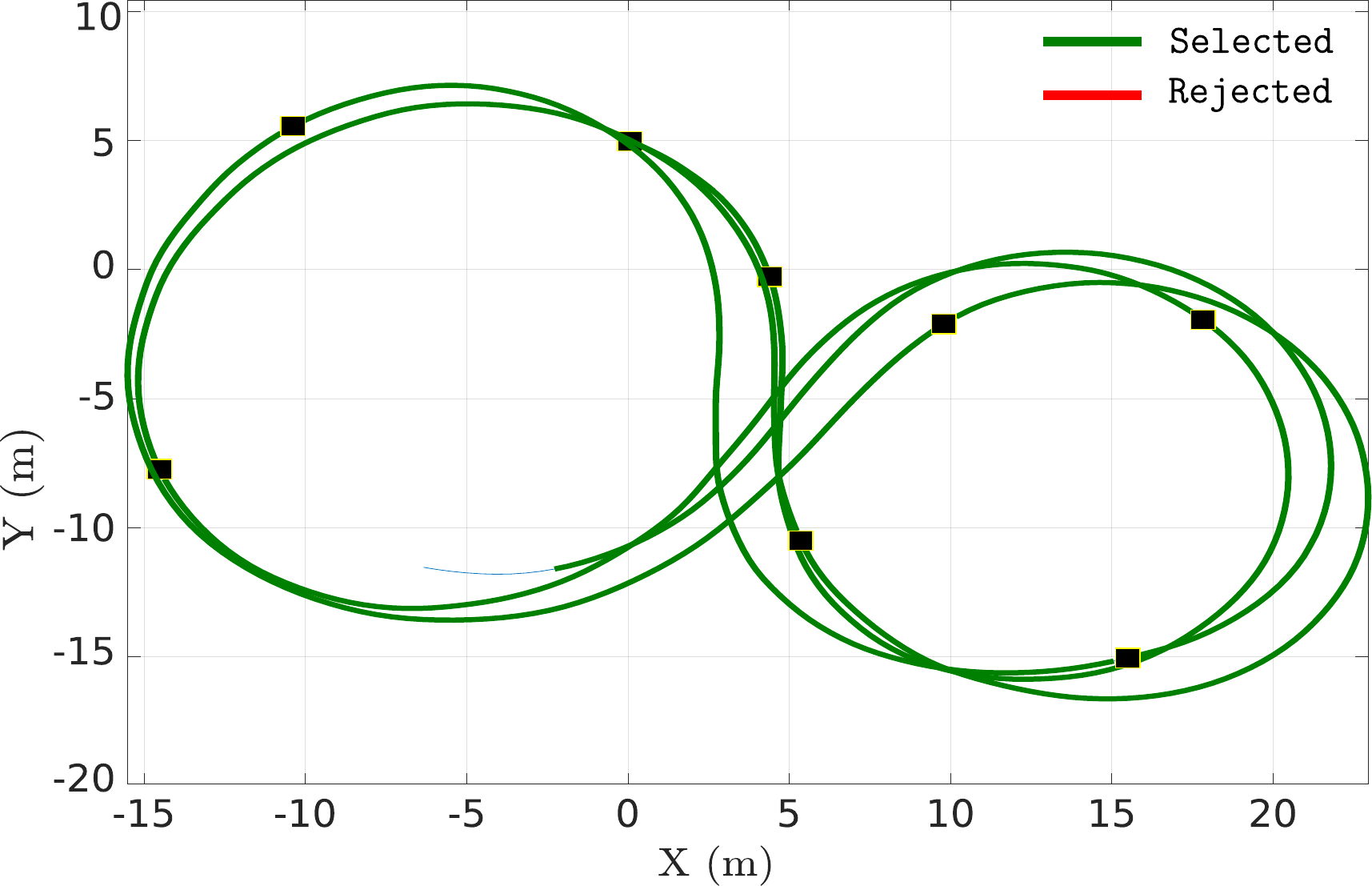}
			%\vspace{-5mm}
			\caption{Selected trajectory segments with observability criteria fulfilled for Seq-1 shown in green color.}
			\label{fig:obs_traj_1}
			%\vspace{-4mm}
		\end{subfigure}
		\hfill
		\begin{subfigure}[b]{1\columnwidth}
			%\vspace{-1mm}
			\centering
			\includegraphics[width=1\columnwidth]{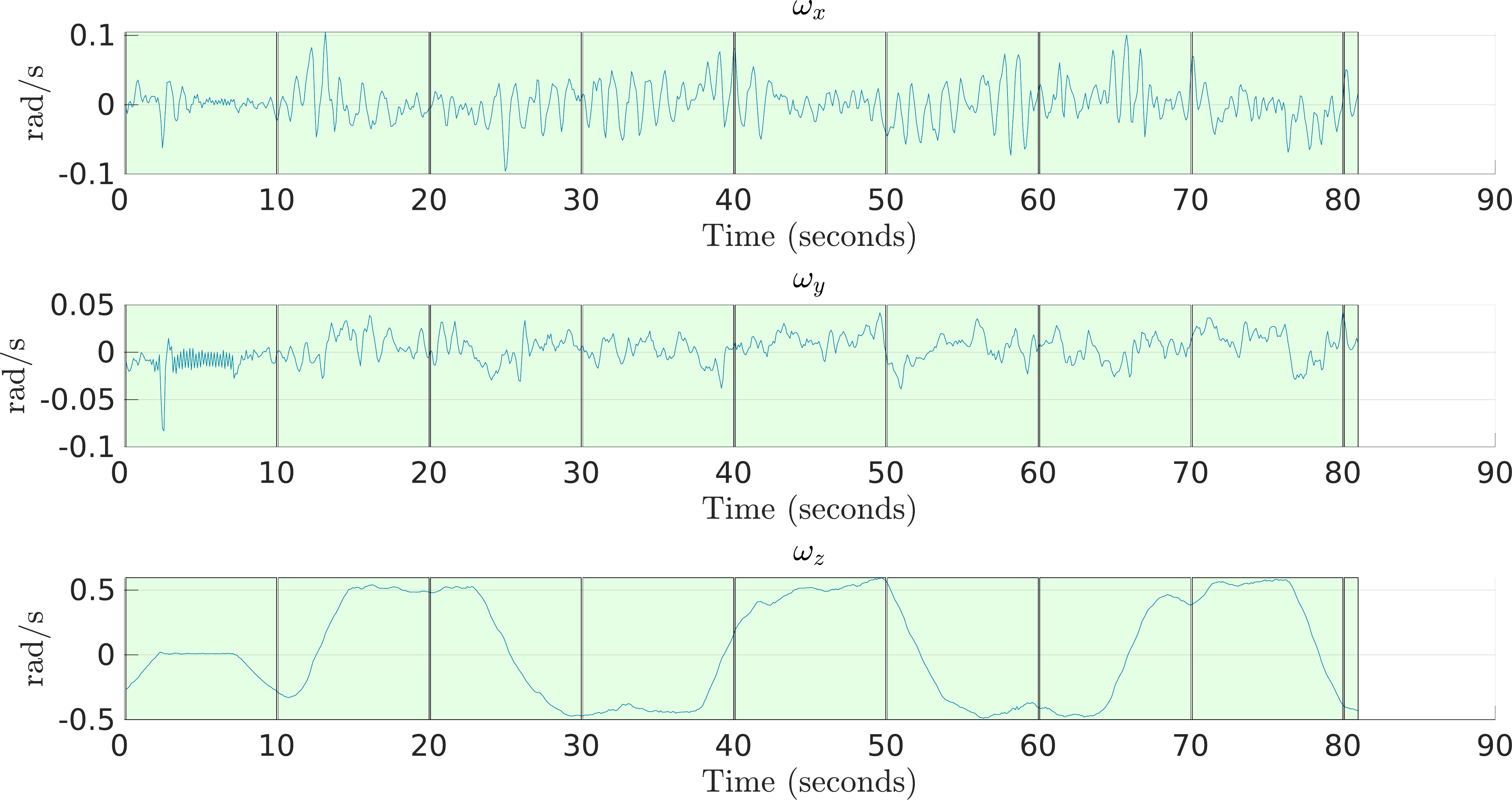}
			\caption{The angular velocity signals in different motion segments in a 10 sec window.}
			\label{fig:obs_ang_vel_1}
		\end{subfigure}
		\caption{Seq-1: The observability criteria selects all the motion primitives as there were enough excitation in the angular velocity signals in all the motion segments.}
		\vspace{-2mm}
	\end{figure}
\begin{figure}
		\centering
		\begin{subfigure}[b]{1\columnwidth}
			\centering
			\includegraphics[width=1\columnwidth]{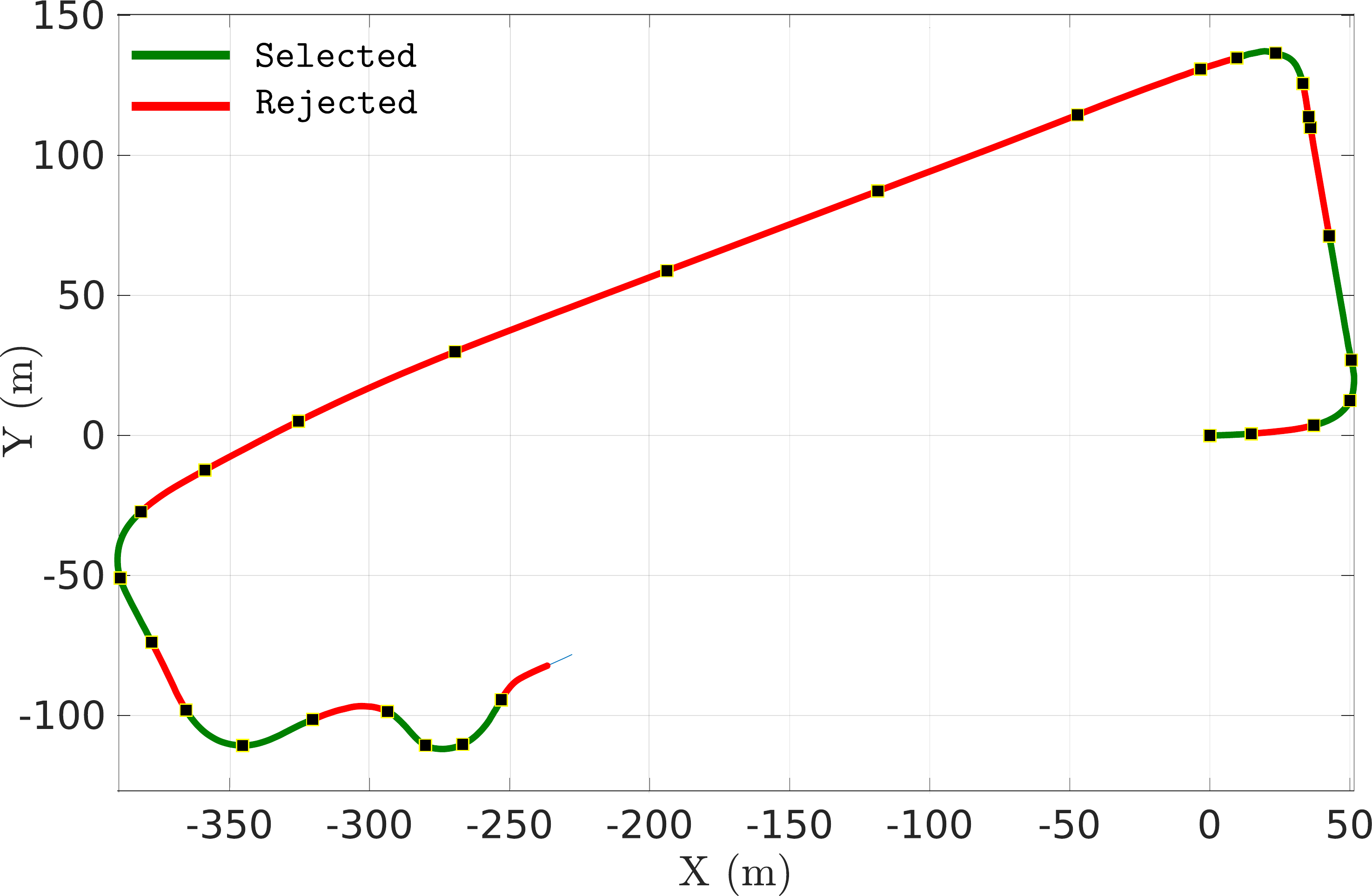}
			%\vspace{-5mm}
			\caption{Selected trajectory segments with observability criteria fulfilled for Seq-4 are shown in green color, whereas rejected trajectory segments are highlighted as red.}
			\label{fig:obs_traj_4}
			%\vspace{-4mm}
		\end{subfigure}
		\hfill
		\begin{subfigure}[b]{1\columnwidth}
			%\vspace{-1mm}
			\centering
			\includegraphics[width=1\columnwidth]{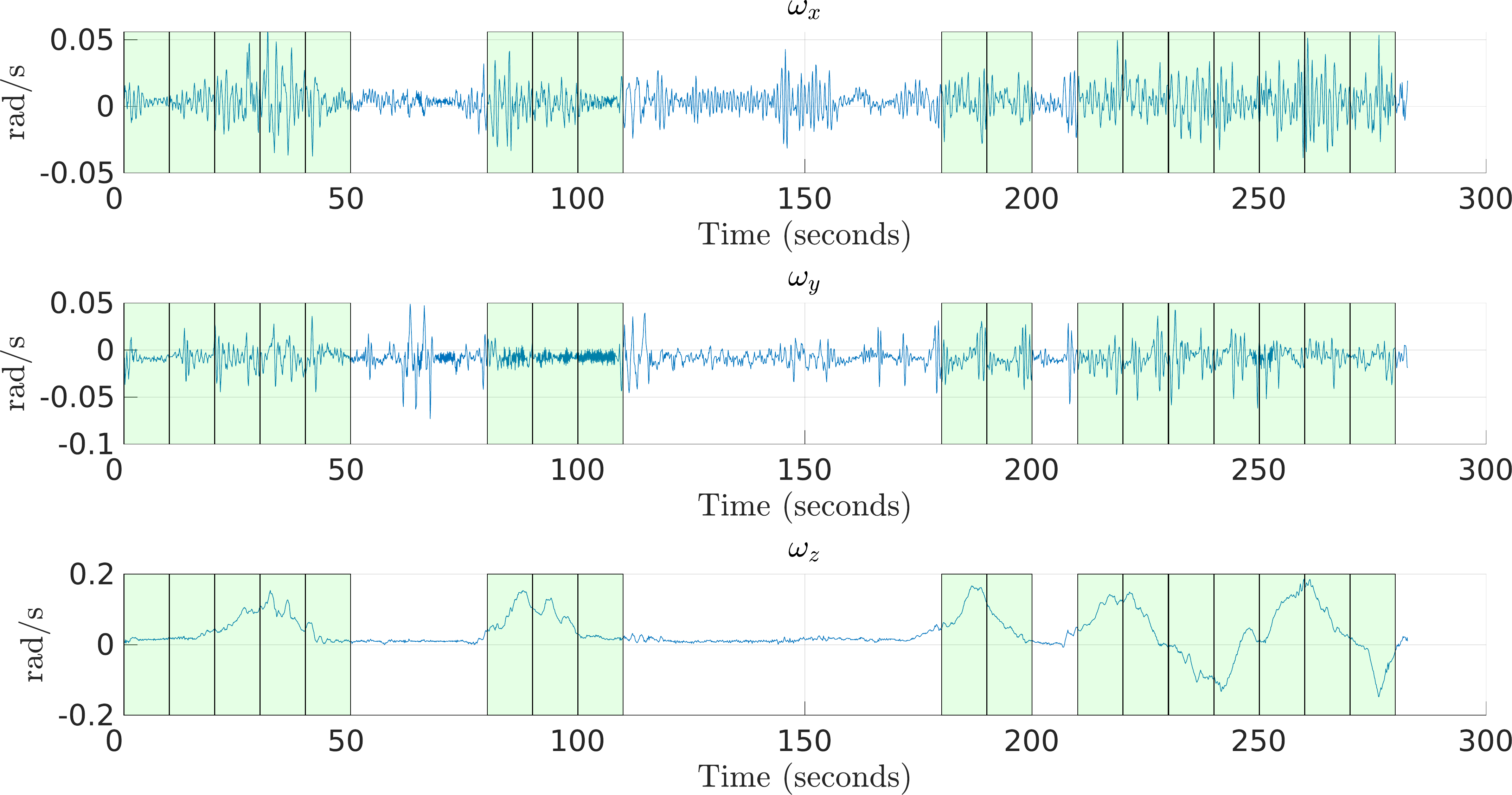}
			%\vspace{-7mm}
			\caption{We observe that in straight motion segments, there is not enough excitation in the angular velocity for Seq-4. Hence, the straight motion segments are not used from an observability perspective and are thus ignored for calibration.}
			\label{fig:obs_ang_vel_4}
		\end{subfigure}
		\caption{Seq-4: The observability criteria selects motion segments when there is enough excitation in the angular velocity signals, particularly during the turns.}
	\end{figure}

	As seen in \Figure\ref{fig:obs_ang_vel_1}, we selected all segments (highlighted in green) as there was enough excitation in the angular velocity during all the time in Seq-1 as the vehicle was driving in a figure-of-8 pattern. The corresponding selected trajectory segments are shown in \Figure\ref{fig:obs_traj_1}. Similarly, for Seq-4, we can see that in \Figure\ref{fig:obs_traj_4}, relevant trajectories for calibration are selected only when there are strong turns maneuvers are observed in \Figure\ref{fig:obs_ang_vel_4} as this is when there is enough excitation of the different degrees of freedom. This analysis helped us to understand that the best possible maneuvers for IMU-based initialization must contain aggressive turns like 8-patterns. 

%	\mfallon{fig 9 is silly if all the sections were selected!}
%	\sandipan{It shows that since we do a lot of turns we get the interesting sections overall. This gives an idea how calibration sequence should be collected! }
	
	\subsection{Online calibration experiments}
	GT calibration of the lidars is obtained by refining the vehicle CAD parameters. The refinement process analyzes known static feature positions around the vehicle in the world frame, $\World$, and matches the corresponding detected features from the lidar perspective, using $\mathbf{T}_{\mathtt{W}\mathtt{B}}$ from GNSS. After that, we obtain the extrinsics of the lidars to the base frame, $\Base$ and compute the transformation matrix, $\mathbf{T}_{\mathtt{G}\mathtt{T}} = \mathbf{T}^{-1}_{\Base\Lidar^{(\mathtt{F}_\mathtt{L}-\mathtt{Top})}} \mathbf{T}_{\Base\Lidar^{(\mathtt{F}_\mathtt{R}-{\mathtt{Top}})}}$, between the 2 lidars. Ideally, GT calibration should be recomputed regularly. However, for our sequences, even though they were collected on different days we used the same GT parameters. 
	
	%Seq-1: 2022-09-16-I
	%Seq-2: 2022-09-16-II
	%Seq-3: 2022-04-07-I
	%Seq-4: 2022-04-07-II
	
	\normalsize
	\begin{table}[!h]
		\centering
		\resizebox{0.99\columnwidth}{!}{
			\begin{tabular}{l|c|cccccc}  \toprule
				\multicolumn{8}{c}{Performance comparison of calibration routines: Scania dataset}\\
				\midrule
				Sequence & Method &
				$\Delta x [m]$  & $\Delta y [m]$ & $\Delta z [m]$ & $\Delta roll [^{\circ}]$ & $\Delta pitch [^{\circ}]$ & $\Delta yaw [^{\circ}]$\\
				\midrule
				\multirow{3}{*}{\raisebox{\heavyrulewidth}{Seq-1}} & Kabsch & 0.367 & \textbf{0.099} & -0.309 & -1.533 & -2.647  & -5.744\\
				& CROON\textsuperscript{\dag} & -0.844 &  -0.959 & -0.340 & -1.707 & -2.783  & 1.918\\
				& Ours$^{\star}$ & \textbf{-0.164}  & -0.184 & \textbf{-0.153} & \textbf{-1.415} & \textbf{-2.759} & \textbf{-0.592}\\
				\midrule
				\multirow{3}{*}{\raisebox{\heavyrulewidth}{Seq-2}} & Kabsch & 0.734 & -0.822 & -0.342 & \textbf{-1.654} & -2.860  & -1.019 \\
				& CROON\textsuperscript{\dag} & \textbf{-0.134} & -0.099  & \textbf{-0.178} & -1.761 & -2.722  &  \textbf{0.265}\\
				& Ours$^{\star}$ & 0.187 & \textbf{0.043} & -0.183 & -2.105 & \textbf{-2.562} & -1.889\\
				\midrule
				\multirow{3}{*}{\raisebox{\heavyrulewidth}{Seq-3}} & Kabsch & 0.675 & -0.392 & 0.440 & -2.474 & -1.940 & \textbf{0.083}\\
				& CROON\textsuperscript{\dag} & 0.310  & \textbf{-0.030}  &  0.081 & -3.387  &  -1.762  & -1.710
				\\
				& Ours$^{\star}$ &  \textbf{0.261} & 0.095 & \textbf{-0.072} & \textbf{0.057} & \textbf{-1.561} & -0.159\\
				\midrule
				\multirow{3}{*}{\raisebox{\heavyrulewidth}{Seq-4}} & Kabsch & 1.260  & -0.339  & 0.870 & 1.784   & -4.947  &  0.304\\
				& CROON\textsuperscript{\dag} & 0.494  & 0.403  & \textbf{0.054} & -1.036 &  -1.447  & -4.654 \\
				& Ours$^{\star}$ & \textbf{0.118} & \textbf{-0.112} & -0.059 & \textbf{0.438}  & \textbf{-1.267} & \textbf{0.138}\\
				\bottomrule
				\multicolumn{8}{l}{\textit{$^{\star}$ Only translation initialization needed}, \textit{\textsuperscript{\dag} Both translation and rotation initialization needed}.}
		\end{tabular}}
		\caption{}
		\label{tab:calib_results_scania_dataset}
		\vspace{-4mm}
	\end{table}
	\begin{figure*}[!hbt]
		\vspace{-3mm}
		\centering
		\includegraphics[width=1\textwidth]{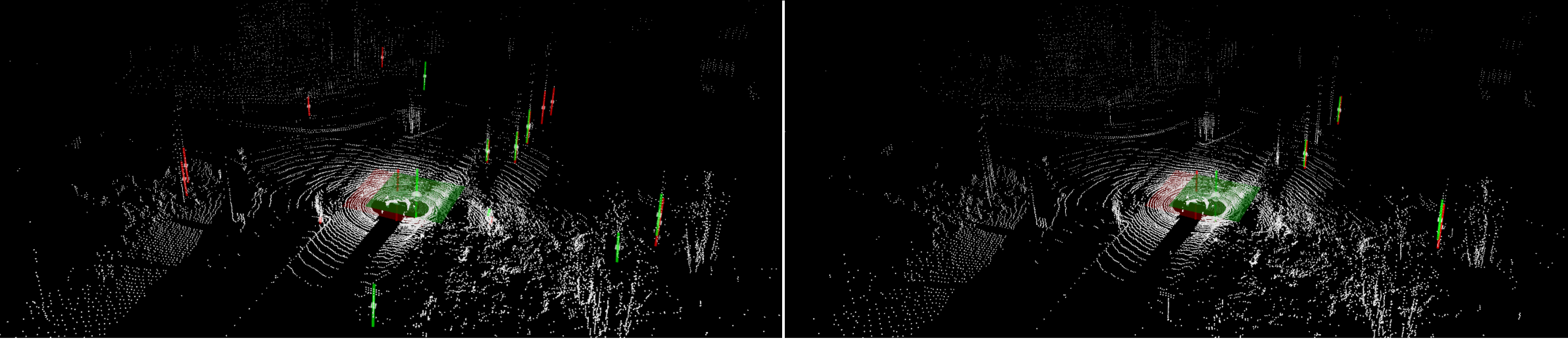}
		\vspace{-5mm}
		\caption{Line and plane features: All (left) and closest (right) features after calibration refinement from Seq-4.}
		\label{fig:features}
		\vspace{-0.7mm}
	\end{figure*}
	We compare the results after evaluating $\mathbf{T}_{\mathtt{G}\mathtt{T}} - \hat{\T}_{\mathtt{Estimated}}$ for all the sequences using Kabsch alignment, CROON and our method. In the refinement step, we match a set of  $N=100$ lines before applying the optimization discussed in Sec.~\ref{sec:refinement}. Kabsch alignment requires lidar odometry and CROON requires ICP matching with normal estimation. Hence for both methods, the computation time is significantly higher. Also, our method does not need any rotation initialization and we can also verify the quality of the extrinsic parameters online which is not considered in these works. As seen in Table~\ref{tab:calib_results_scania_dataset}, our method performs better in feature rich environments present in Seq-3 and Seq-4 and under aggressive maneuvers in Seq-1, where rich motion sequences are available. In \Figure\ref{fig:features} once can see the closest features from the two point clouds after refinement. The features extracted from $\mathtt{F_L-Top}$ and $\mathtt{F_R-Top}$ lidar are colored red and green respectively. 

	\section{Conclusion}
	\label{sec:conclusion}
	In this work, we show that it is possible to initialize the extrinsic calibration of multiple lidars by matching the signals from IMUs co-located with the lidars. Unlike other methods which use odometry estimates for matching poses, this is a lightweight method that relies on raw signal matching. An observability-aware module informs us of the maneuvers needed to produce the excitations necessary for successful extrinsic calibration. %\mfallon{This doesnt make sense. A method cant be comparable to GT. Ground truth is the true answer you cannot be comparable to the true answer. Perhaps you mean that `the best estimate from the vehicle's cad'?} 
	Our method provides comparable extrinsics to the vehicle's CAD parameters when sufficient signal excitations are present in normal urban driving scenarios. %\mfallon{why this? without any special setup}. 
	
	\section{Acknowledgements}
	This research has been jointly funded by the Swedish Foundation for Strategic Research (SSF) and Scania. The research is also affiliated with Wallenberg AI, Autonomous Systems and Software Program (WASP).
	
	%\addtolength{\textheight}{-1cm}  % This command serves to balance the column
	% lengths on the last page of the document manually. It shortens the textheight
	% of the last page by a suitable amount. This command does not take effect
	%until
	% the next page so it should come on the page before the last. Make sure that
	% you do not shorten the textheight too much.
	
	%%%%%%%%%%%%%%%%%%%%%%%%%%%%%%%%%%%%%%%%%%%%%%%%%%%%%%%%%%%%%%%%%%%%%%%%%%%%%%%%
	
	\bibliographystyle{./IEEEtran}
	\bibliography{./IEEEabrv, ./library}
	
	%%Bibtex cleaner: https://flamingtempura.github.io/bibtex-tidy/
	
\end{document}